\documentclass[journal]{IEEEtran}
\ifCLASSINFOpdf
\else
\fi

\usepackage{graphicx}%
\usepackage{multirow}%
\usepackage{amsmath,amssymb,amsfonts}%
\usepackage{amsthm}%
\usepackage{mathrsfs}%
\usepackage{xcolor}%
\usepackage{textcomp}%
\usepackage{manyfoot}%
\usepackage{booktabs}%
\usepackage{algorithm}%
\usepackage{algorithmicx}%
\usepackage{algpseudocode}%
\usepackage{listings}%

\usepackage{caption}
\usepackage{subcaption}
\usepackage{standalone}
\usepackage{tikz}
\usetikzlibrary{positioning, matrix, fit, calc}
\usepackage{pgfpages}%

\usepackage[hidelinks]{hyperref}  

\usepackage{enumitem}

\usepackage{comment}

\usepackage{soul}

\usepackage{nomencl}
\makenomenclature
\newtheorem{theorem}{Theorem}
\newtheorem{proposition}[theorem]{Proposition}%

\begin{document}

\title{Budget-constrained Collaborative Renewable Energy Forecasting Market}

\author{Carla Gon\c{c}alves, Ricardo J. Bessa,~\IEEEmembership{Senior~Member,~IEEE}, Tiago Teixeira, Jo\~{a}o Vinagre
\thanks{C. Gon\c{c}alves and R. J. Bessa are with INESC TEC, Campus da FEUP, 4200-465 Porto, Portugal (e-mail: \{carla.s.goncalves, ricardo.j.bessa\}@inesctec.pt). T. Teixeira was with the Faculty of Sciences of the University of Porto (FCUP), Portugal. 
J. Vinagre is with the Joint Research Centre (European Commission) and the European Centre for Algorithmic Transparency (ECAT), Seville, Spain. 
}
\thanks{The research leading to this work is being carried out as a part of the ENERSHARE (\textit{European commoN EneRgy dataSpace framework enabling data sHaring-driven Across- and beyond- eneRgy sErvices}) project, European Union’s Horizon Research and Innovation Programme, Grant Agreement No. 101069831. Views and opinions expressed are however those of the author(s) only and do not necessarily reflect those of the European Union. Neither the European Union nor the granting authority can be held responsible for them.}
}

\maketitle

\begin{abstract}
Accurate power forecasting from renewable energy sources (RES) is crucial for integrating additional RES capacity into the power system and realizing sustainability goals. This work emphasizes the importance of integrating decentralized spatio-temporal data into forecasting models. However, decentralized data ownership presents a critical obstacle to the success of such spatio-temporal models, and incentive mechanisms to foster data-sharing need to be considered. The main contributions are {a)}~a comparative analysis of the forecasting models, advocating for efficient and interpretable spline LASSO regression models, and {b)}~a bidding mechanism within the data/analytics market to ensure fair compensation for data providers and enable both buyers and sellers to express their data price requirements. Furthermore, an incentive mechanism for time series forecasting is proposed, effectively incorporating price constraints and preventing redundant feature allocation.    
Results show significant accuracy improvements and potential monetary gains for data sellers. 
For wind power data, an average root mean squared error improvement of over 10\% was achieved by comparing forecasts generated by the proposal with locally generated ones.
\end{abstract}

%
%
\nomenclature{AR}{AutoRegressive}
\nomenclature{ARMA}{AutoRegressive Moving Average}
\nomenclature{ARIMA}{AutoRegressive Integrated Moving Average}
\nomenclature{LASSO}{Least Absolute Shrinkage and Selection Operator}
\nomenclature{LRM}{LASSO Regression analytics Market}
\nomenclature{MSE}{Mean Squared Error}
\nomenclature{RES}{Renewable Energy Sources}
\nomenclature{RMSE}{Root Mean Squared Error}
\nomenclature{SLCM}{Spline LASSO bid-Constrained Market}
\nomenclature{ZRM}{Zero-Regret auction Mechanism}
%
\nomenclature{$(x)_+$}{Rectified linear unit function, $(x)_+ = \max(x,0)$}
\nomenclature{$\mathcal A$}{Set of all data owners, $\mathcal A=\{1, \dots, N\}$}
\nomenclature{$\mathcal B$, $\mathcal S$}{Set of data buyers ($\mathcal B \subseteq \mathcal A$) and sellers ($\mathcal S \subseteq \mathcal A$), resp.}
\nomenclature{$N$}{Number of data owners}
\nomenclature{$n_i$}{Number of exogenous variables from agent $i$}
\nomenclature{$T$}{Number of available timestamps}
\nomenclature{$H$}{Prediction horizon}
\nomenclature{$Y_{i,t}$}{Target variable for agent $i$ at time $t$}
\nomenclature{$\lambda$}{LASSO penalty term}
\nomenclature{$D, K$}{Spline transformer degree and number of knots, resp.}
\nomenclature{$X_{i,t}^{k}$}{$k$th feature from $i$th agent at time $t$}
\nomenclature{$X_{i,m,t}^{k}$}{$m$th spline of $k$th feature from $i$th agent at time $t$}
\nomenclature{$\mathcal X_{i,t}$}{Features of agent $i$ at time $t$}
\nomenclature{$\mathcal Z_{t}$}{Available features at time $t$}
\nomenclature{$\mathbf X_i$, $\mathbf Z$}{Matrix of all historical \(\mathcal X_{i,t}\) and \(\mathcal Z_{t}\) observations}
\nomenclature{$\mathbf y_i$}{Vector of all historical \(Y_{i,t}\) observations}
\nomenclature{$\beta^i_{j,k}$}{Coefficient for \(k\)th feature of agent \(j\) predicting \(Y_{i,t}\)}
\nomenclature{$\beta_{j,k,m}^i$}{Coefficient for $X_{j,m,t}^{k}$ predicting $Y_{i,t}$}
\nomenclature{$\eta_{j,\ell,m}^i$}{Coefficient for $m$th spline of $Y_{j-t-\ell}$ predicting $Y_{i,t}$}
\nomenclature{$\boldsymbol{\Theta}$}{Parameters of the forecasting models}
\nomenclature{$\mathcal G$}{Gain with collaborative model, e.g., loss improvement}

\nomenclature{$f_\text{local}^{i}$, $f_\text{market}^{i}$}{Local and collaborative models, respectively}
\nomenclature{$\mathcal L(\cdot)$}{Loss function}
\nomenclature{$\mathcal L_\text{local/market}^{i}$}{Local and collaborative loss functions, resp.}
\nomenclature{$\psi_j$}{Relative contribution from feature $j$ (Shapley value)}
\nomenclature{$\mathcal{VF}$}{Function that relates gain with bids}
\nomenclature{$\mathcal{I}(x)$}{Indicator function, $\mathcal{I}(x){=}\{1 \text{ if } x{\neq} 0, 0 \text{ otherwise}\}$}
\nomenclature{$\mathcal{ST}$}{Spline transformer}
\nomenclature{$\mathcal{FS}$}{Filter selection: partial Pearson correlation test}
\nomenclature{$r_i$}{Payment to be made by $i$th data buyer}
\nomenclature{$b_i$}{Bid price from $i$th data buyer}
\nomenclature{$p^*_i$}{Market price for the $i$th data buyer}
\nomenclature{$H_j^k(\cdot)$}{Payment for the $k$th feature of agent $j \in \mathcal{S}$}
\nomenclature{$u_{j,k}$}{Agent $j$'s reservation to sell $k$th feature}
\nomenclature{$s_{j,k}$}{Agent $j$'s price to the $k$th feature, \(\boldsymbol{s}_j {=} (s_{j,1}, {\allowbreak} ..., s_{j,n_j})\)}
\printnomenclature

\section{Introduction}

\IEEEPARstart{R}{enewable} energy sources (RES) are critical in addressing the decarbonization of the economy. However, integrating RES into the power grid efficiently poses challenges owing to their variable and uncertain nature, given their dependence on ever-shifting weather conditions. Accurate RES forecasting is vital to ensure its reliability and optimize RES integration from the technical and economic viewpoint, leading to enhanced efficiency and cost reduction. Many studies have shown that combining data from different geographical locations can significantly improve forecasting models by capturing spatio-temporal dependencies~\cite{Tilmann2006, cavalcante2017lasso}.  These dependencies are a direct outcome of weather patterns, where conditions at a specific location and time are intricately linked to their prior states and adjacent locations. Although recognizing such dependencies is not novel, as evidenced by the work of Tilmann et al.~in 2006~\cite{Tilmann2006}, a notable barrier to promoting effective data combinations lies in the decentralized ownership of data. Multiple entities maintain their databases, often reluctant to share data for competitive reasons. 

To tackle this barrier, the European Commission created a set of harmonized rules for fair access to and use of data, in a regulation known as the Data Act, which entered into force in January 2024. This strategy considers the Common European Data Spaces concept that aims to ensure that more data becomes available in the economy, society, and research while keeping the companies and individuals in control of the data they generate~\cite{Otto2022}. In this context, establishing incentive mechanisms for data-sharing is crucial. Privacy-preserving techniques, like federated learning, create technical conditions for collaborative analytics~\cite{Li2021}. For instance, training a vector autoregressive model that combines data from multiple parties without revealing private information~\cite{goncalves2021privacy}. However, relying solely on data privacy might not be sufficient. For example, data owner A can greatly improve accuracy when using data from data owner B, while data owner B may not have any benefit when using data from A. Although data privacy is ensured, data owner B has no cooperation incentive.

The emergence of data and analytics markets focused on monetizing data or related statistics~\cite{PARRAARNAU2018354, mehta2021sell} is an incentive mechanism to tackle this issue. Potential data buyers include, for instance, a) transmission system operators seeking real-time data from local weather stations or forecasts from various vendors, with compensation based on their specific contributions to improving particular objectives (e.g., reducing balancing costs through enhanced operational planning); b) wind power plant operators and/or balancing responsible parties aiming to enhance predictability to lower imbalance penalties or optimize the operation of local energy storage devices; c) distribution system operators, who could benefit from data on groups of consumers connected to the same primary or secondary substation, particularly recent measurements from those with electric vehicles, storage systems, or smart appliances. A detailed description of different business models and interested parties is presented in~\cite{Wang2023}.

A fundamental challenge in these markets is determining the value of data and establishing fair pricing mechanisms. The work in~\cite{agarwal2019marketplace} pioneered constructing a data market where buyers purchase forecasts instead of specific datasets based on cooperative game theory. Relevant features of this framework include equitable revenue distribution among sellers, payment based on improved forecasting skills, and compensation for incremental gain. Based on this previous work, an adapted framework for electricity markets is presented in~\cite{carla2021}, which considers continuous updating of the input variables, incorporates lagged time-series data, and establishes a relationship between the exchanged data and reduced imbalance costs. However, this approach has limitations, including the failure to consider sellers' perspectives and the computational complexity involved in data monetization due to, e.g., the usage of Shapley values (making it unsuitable for intraday markets). Other methods set prices based on seller bids, as in~\cite{trading22}. A recurring gap in the literature is the predominant influence of one party on price determination.

These limitations highlight the need to refine data market design, ensure equilibrium between buyer and seller interests, and improve marketplace capabilities for rapid forecasting and real-time decisions.

The contributions of this paper are summarized as follows:
\begin{enumerate}
\item A novel data bidding and pricing mechanism that allows sellers to set the exact prices for their data within a collaborative forecasting model. Simultaneously, buyers can set a maximum price or a price based on the accuracy improvement achieved through collaborative forecasting, compared to relying solely on their local data.

\item An interpretable collaborative forecasting model (spline LASSO), where the coefficients, payments, and revenues are automatically determined by solving a sequence of cost-constrained optimization problems.
\end{enumerate}

The rest of this paper is structured as follows. Section~\ref{sec:literature} formulates the RES power forecasting problem and reviews related work on forecasting and monetary incentives for data-sharing. Section~\ref{sec:proposal} presents the novel data market formulation for regression tasks. Numerical results and comparisons are made in Section~\ref{CaseStudy}. Possible extensions are discussed in Section~\ref{sec:classification}, and the conclusions presented in Section~\ref{sec:conclusion}.


\section{Problem Formulation and Related Work}\label{sec:literature}

The scope of this paper is a monetary data-sharing incentive mechanism applied to a day-ahead forecasting problem involving multiple RES power plants. Henceforth, Section~\ref{notation} introduces the notation and formulates the forecasting problem, Section~\ref{models} reviews the key forecasting approaches, and Section~\ref{datamarkets} reviews the most relevant algorithmic solutions for monetary data-sharing incentives.

\subsection{Formulation of the RES Forecasting Problem} \label{notation}

Consider a set of $N$ agents, denoted as $\mathcal{A}=\left\{1, 2, \ldots, N\right\}$, representing the owners of the RES power plants. For simplicity, consider also that agent $i$ operates a single power plant, having a single target variable $Y_{i, t}$, corresponding to the produced power. Additionally, each agent observes a set of $n_i$ exogenous variables, such as wind speed or irradiance forecasts, denoted by $\mathbf X_{i,t}=\left\{X_{i, t}^1, \ldots, X_{i, t}^{n_i}\right\}$. Here, $X_{i, t}^j$ denotes the $j$th variable from the $i$th agent, $i \in \mathcal{A}$.

At time $t_0$, the goal of agent $i$ is to forecast $\left\{Y_{i,t}\right\}_{t=t_0+1}^{t_0+H}$, where $H$ is the number of look-ahead timestamps. Two models can be considered depending on the available data:

\begin{itemize}
\item \textbf{Local model}, where agent $i$ only has local data, a model function $f(\cdot)$ can be used for each horizon such that
\begin{equation}
Y_{i,t} \approx f_{\text{local}}^{i}(\mathcal X_{i, t}; \boldsymbol \Theta_i),
\end{equation}
where $\boldsymbol\Theta_i$ are the parameters to be estimated and
\begin{equation}
\mathcal{X}_{i,t} = \underbrace{\{X_{i, t}^1, \ldots, X_{i, t}^{n_i}}_{\text{exogenous from agent i}}, \underbrace{Y_{i,t_0-1}, \ldots, Y_{i,t_0-L}\}}_\text{L most recent power meas.}.
\end{equation} 

\item \textbf{Collaborative model}, where agents share their data,
\begin{equation}
Y_{i,t} \approx f_{\text{market}}^{i}(\mathcal Z_{t}; \boldsymbol\Theta_i),
\end{equation}
where $\mathcal Z_{t} = \{\mathcal X_{1,t}, \mathcal X_{2,t}, \ldots, \mathcal X_{N,t}\}$.
\end{itemize}

The mapping function $f(\cdot)$ can exhibit either linear or nonlinear characteristics, depending on the forecasting horizon, and a multitude of approaches can be explored to model this function effectively. The next section presents a comprehensive discussion of the approaches to model $f(\cdot)$.

\subsection{RES Forecasting Models}\label{models}

The RES forecasting literature over the last years has focused on different variants of statistical and machine learning approaches~\cite{tawn2022review}. Classical time series models, including autoregressive (AR)~\cite{miranda2006one}, autoregressive moving average (ARMA)~\cite{rajagopalan2009wind}, and autoregressive integrated moving average (ARIMA)~\cite{reikard2009predicting}, have been widely used for forecasting a few hours-ahead of individual time series. Extensions of these models, such as vector autoregressive (VAR)~\cite{cavalcante2017lasso, cavalcante2017solar}, have proven effective for multivariate very short-term (e.g., up to 6 hours ahead) RES forecasting, capturing spatio-temporal dependencies. These models have also integrated exogenous variables, extending AR to ARX and VAR to VARX.

In statistical learning, linear regression serves as a fundamental and commonly used forecasting model, explaining the expected value of a variable with a linear combination of covariates. 
However, high covariate correlation can complicate interpretation and lead to overfitting. To address this, regularization terms have been added to the loss function to promote sparsity in coefficient estimation and facilitate automatic feature selection. 
While regularized linear regression models offer advantages, they struggle with capturing nonlinear relationships between variables, as is the case between wind power and speed. To address this, researchers have turned to linear additive models~\cite{Nielsen2006}, which provide a flexible framework for complex relationships. One approach involves transforming variables to enable linear models in the coefficients, such as spline regression~\cite{tastu2013probabilistic} and kernelized linear regression~\cite{naik2018short}. In addition to these, various non-linear models have been used for RES forecasting, including support vector machines, multi-layer perceptron~\cite{muhammad2017day}, random forests~\cite{lahouar2017hour}, gradient-boosting regression trees~\cite{andrade2017improving}, and ensembles of these techniques. Deep neural networks have also been explored~\cite{ying2022deep}, with architectures combining, for example, convolution neural networks with long short-term memory networks to capture spatial and temporal dependencies.

\subsection{Data Markets}\label{datamarkets}

\begin{figure*}
\begin{subfigure}{0.325\textwidth}
\begin{tikzpicture}[font=\small]
\matrix (text) [inner sep=2pt, matrix of nodes, column 2/.style={nodes={align=left, text width=2.95cm}}] {
& \bf Market Operator \\
\bf S1 & Agents send $\{\mathcal X_{j,t}\}$ \\
\bf S2 & Market fixes $p^*_i$ \\
\bf S3 & Buyer sets $f_i, \mathcal G_i, b_i$ \\
\bf S4 & Noise addition if $b_i{<}p^*_i$, $f_i$ estimation \\
\bf S5 & Payment $p_i$ ($\mathcal G_i$-based) \\
\bf S6 & Revenue $r_j$ (Shapley) \\
};

\coordinate (bottom) at ($(text.north) - (0, 3.6cm)$);
\node[inner sep=0pt, fit=(text)(bottom), draw] {};

\node[inner xsep=0pt, xshift=-4mm, anchor=north east] at (text.north west) (buyer) {Buyer $i$};

\draw[-latex, shorten >=1mm] ([xshift=3mm]buyer.south west) |- node [above, pos=0.75, font=\scriptsize, align=left] {$\{Y_{i,t}\}_{\tiny 1}^{\tiny T}$\\$f_i$,$\mathcal{G}_i$,$b_i$} (text-4-1);

\draw[-latex, shorten >=1mm] ([xshift=2mm]buyer.south west) |- node [pos=0.75, above, font=\scriptsize] {$p_i$} ([yshift=3mm]text-6-1);

\draw[latex-, shorten >=1mm] ([xshift=1mm]buyer.south west) |- node [pos=0.72, below, font=\scriptsize] {$\{\hat Y_{i,t}\}_{T{+}1}^{T{+}H}$} ([yshift=-3mm]text-6-1);

\matrix (r) [inner sep=0pt, matrix of nodes, row sep=5pt] at ([xshift=3mm]text-2-2.east) {
$1$ \\
$\dots$ \\
$N$ \\
};

\draw[-latex, shorten <=0.8mm] (text-7-2.east) -- ++(8mm,0) |- node[above, pos=0.75, font=\scriptsize] {$r_1$} (r-1-1);
\draw[-latex, shorten <=0.8mm] (text-7-2.east) -- ++(8mm,0) |- node[above, pos=0.75, font=\scriptsize] {$r_N$} (r-3-1);

\draw[-latex] (r-1-1) -- ++(-4mm,0) |- ([xshift=-4mm]text-2-2.east);
\draw[-latex] (r-3-1) -- ++(-4mm,0) |- ([xshift=-4mm]text-2-2.east);
\end{tikzpicture}%
\caption{Cooperative zero-regret~\cite{carla2021}.}
\label{fig:market1}
\end{subfigure}%
\hfill
\begin{subfigure}{0.325\textwidth}
\begin{tikzpicture}[font=\small]

\matrix (text) [inner sep=2pt, matrix of nodes, column 2/.style={nodes={align=left, text width=2.95cm}}] {
& \bf Market Operator \\
\bf S1 & Agents send $\{\mathcal X_{j,t}\}$ \\
\bf S2 & Buyer sets $f_i, \mathcal L_i, b_i$ \\
\bf S3 & Update $f_i$ coefficients (Newton-Raphson) \\
\bf S4 & Payment $p_i{=}\allowbreak(\mathcal L_{local}{-}\mathcal L_{market}){\times} b_i$ \\
\bf S5 & $r_j {=} p_i \cdot \sum_{k \in \mathcal{X}_{j,t}} \psi_{j,k}$ \\
};

\coordinate (bottom) at ($(text.north) - (0, 3.6cm)$);
\node[inner sep=0pt, fit=(text)(bottom), draw] {};

\node[inner xsep=0pt, xshift=-4mm, anchor=north east] at (text.north west) (buyer) {Buyer $i$};

\draw[-latex, shorten >=1mm] ([xshift=3mm]buyer.south west) |- node [above, pos=0.75, font=\scriptsize, align=left] {$\{Y_{i,t}\}_{\tiny 1}^{\tiny T}$\\$f_i$,$\mathcal{L}_i$,$b_i$} (text-3-1);

\draw[-latex, shorten >=1mm] ([xshift=2mm]buyer.south west) |- node [pos=0.75, above, font=\scriptsize] {$p_i$} ([yshift=3mm]text-5-1);

\draw[latex-, shorten >=1mm] ([xshift=1mm]buyer.south west) |- node [pos=0.72, below, font=\scriptsize] {$\{\hat Y_{i,t}\}_{T{+}1}^{T{+}H}$} ([yshift=-3mm]text-5-1);

\matrix (r) [inner sep=0pt, matrix of nodes, row sep=5pt] at ([xshift=3mm]text-2-2.east) {
$1$ \\
$\dots$ \\
$N$ \\
};

\draw[-latex, shorten <=0.8mm] (text-7-2.east) -- ++(8mm,0) |- node[above, pos=0.75, font=\scriptsize] {$r_1$} (r-1-1);
\draw[-latex, shorten <=0.8mm] (text-7-2.east) -- ++(8mm,0) |- node[above, pos=0.75, font=\scriptsize] {$r_N$} (r-3-1);

\draw[-latex] (r-1-1) -- ++(-4mm,0) |- ([xshift=-4mm]text-2-2.east);
\draw[-latex] (r-3-1) -- ++(-4mm,0) |- ([xshift=-4mm]text-2-2.east);
\end{tikzpicture}%
\caption{Time-adaptive regression~\cite{pierre2022regression}.}
\label{fig:market2}
\end{subfigure}%
\hfill 
\begin{subfigure}{0.325\textwidth}
\begin{tikzpicture}[font=\small]

\matrix (text) [inner sep=2pt, matrix of nodes, column 2/.style={nodes={align=left, text width=2.95cm}}] {
& \bf Market Operator \\
\bf S1 & Agents $\{\mathcal X_{j,t}\}, u_{j,k}$ \\
\bf S2 & Buyer sets $\mathcal L_i$ \\
\bf S3 & $\beta_{j,k}^i$ estimation \eqref{eq:lasso-market} \\
\bf S4 & Payment $p_i{=}\allowbreak\sum_{j\neq i}\sum_k |u_{j,k}\beta_{j,k}^i|$ \\
\bf S5 & $r_j=\sum_k |u_{j,k}\beta_{j,k}^i|$ \\
};

\coordinate (bottom) at ($(text.north) - (0, 3.6cm)$);
\node[inner sep=0pt, fit=(text)(bottom), draw] {};

\node[inner xsep=0pt, xshift=-4mm, anchor=north east] at (text.north west) (buyer) {Buyer $i$};

\draw[-latex, shorten >=1mm] ([xshift=3mm]buyer.south west) |- node [above, pos=0.75, font=\scriptsize, align=left] {$\{Y_{i,t}\}_{\tiny 1}^{\tiny T}$\\$\mathcal{L}_i$} (text-3-1);

\draw[-latex, shorten >=1mm] ([xshift=2mm]buyer.south west) |- node [pos=0.75, above, font=\scriptsize] {$p_i$} ([yshift=3mm]text-5-1);

\draw[latex-, shorten >=1mm] ([xshift=1mm]buyer.south west) |- node [pos=0.72, below, font=\scriptsize] {$\{\hat Y_{i,t}\}_{T{+}1}^{T{+}H}$} ([yshift=-3mm]text-5-1);

\matrix (r) [inner sep=0pt, matrix of nodes, row sep=5pt] at ([xshift=3mm]text-2-2.east) {
$1$ \\
$\dots$ \\
$N$ \\
};

\draw[-latex, shorten <=0.8mm] (text-7-2.east) -- ++(8mm,0) |- node[above, pos=0.75, font=\scriptsize] {$r_1$} (r-1-1);
\draw[-latex, shorten <=0.8mm] (text-7-2.east) -- ++(8mm,0) |- node[above, pos=0.75, font=\scriptsize] {$r_N$} (r-3-1);

\draw[-latex] (r-1-1) -- ++(-4mm,0) |- ([xshift=-4mm]text-2-2.east);
\draw[-latex] (r-3-1) -- ++(-4mm,0) |- ([xshift=-4mm]text-2-2.east);
\end{tikzpicture}%
\caption{LASSO regression~\cite{trading22}.}
\label{fig:market3}
\end{subfigure}
\caption{Related existing algorithmic solutions for analytics trading.}
\label{fig:windfarmcomparison}
\end{figure*}
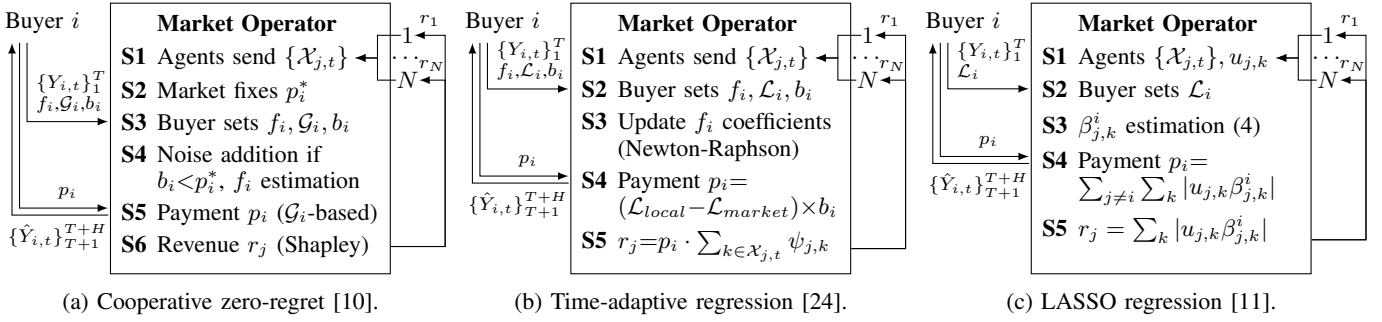  

Data markets serve as platforms facilitating the exchange, acquisition, and sharing of data among multiple stakeholders, ensuring their collective benefit. When combined with collaborative analytics, data markets can be called analytics markets~\cite{falconer2023bayesian}.
These markets typically comprise sellers, buyers, and a central market operator responsible for data storage, security, and overall market functioning. Due to the exclusive access granted to the market operator, concerns related to data confidentiality are addressed and mitigated. Several proposals have been put forth in the literature under this framework. Some assume data owners observe the same features for different instances, and the data price is directly related to the quantity of instances~\cite{cao2017data}. However, this is not the case with time series forecasting, where data owners observe different time series, and buying more time series is not a guarantee of better forecasting accuracy since some may be redundant or non-related to the target. A description of recent solutions for trading time series data is presented next.

The work in~\cite{carla2021} extends the \textit{cooperative Zero-Regret auction Mechanism (ZRM)} in~\cite{agarwal2019marketplace} to incentivize collaborative forecasting in the RES sector. Three main types of agents are considered: data buyers, data sellers, and the market operator. The algorithm is illustrated in Fig.~\ref{fig:market1} and operates as follows. At a specific time $T$, a new session begins, where agents submit their historical data $\left\{\mathcal{X}_{j,t}\right\}_{t=1}^{T+H}$ to the market operator. The market starts by computing a market price $p^*_i \in \mathbb{R}^+$ representing the price per unit increase in forecasting accuracy for this session -- the computation of $p^*_i$  is iterative and based on previous prices and buyers gains.
Upon arrival, agent $i$ requests a power forecast for the next $H$ time steps and submits a bid ($b_i$) indicating the amount he is willing to pay per unit of improvement in forecasting accuracy. Considering the bid ($b_i$) and market price ($p^*_i$), the market operator allocates the available features -- this allocation determines how much noise will be introduced to the data provided by the sellers, in case $b_i<p^*_i$.
Subsequently, the market operator applies cross-validation to estimate the expected gain in forecasting accuracy using the collaborative forecast instead of a local one. The final payment ($p_i$) is computed accordingly and distributed by the sellers according to the importance of their data, assessed by a Shapley-based algorithm. Finally, the buyer $i$ receives the forecasts for $\left\{Y_{i,t}\right\}_{t=T+1}^{T+H}$ and leaves the market session. Throughout the process, the sellers continuously update their data as new time steps occur, ensuring the availability of up-to-date information.

Similarly, the work in~\cite{pierre2022regression} assumes three main entities and considers data sellers sharing their data and accepting the revenue generated by the market operator without establishing any acceptable revenue (zero-regret). However, the market price depends solely on the buyer's bid, and the models can be updated as time progresses instead of re-training the models from scratch as in~\cite{carla2021}. The proposal is depicted in Fig.~\ref{fig:market2}. A regression task is requested by buyer $i$ to the market operator and can be performed either as a batch learning task or an online learning task by a regression model $f_i(\cdot)$ specified by $i$. Buyer $i$ also expresses its willingness to pay, $b_i$, for improving the model accuracy, as measured by a loss function $\mathcal L(\cdot)$. The payment for the $i$th buyer is $p_i = (\mathcal L_\text{local}- \mathcal L_\text{market}) \times b_i, $ 
where the loss values, $\mathcal L_\text{local}$ and $\mathcal L_\text{market}$, are obtained when forecasting $Y_{i,t}$ using $\mathcal X_{i,t}$ and $\mathcal Z_t$, respectively. The bid $b_i$ is stated in monetary terms per unit improvement in the loss function $\mathcal L(\cdot)$ and per data point provided. Data sellers $j \in \mathcal{S}$, $j \neq i$, provide new data as time progresses, allowing the platform to learn and update the regression model continuously by Newton-Raphson method, and consequently $\mathcal L_\text{local}$ and $\mathcal L_\text{market}$. The improvement in the collaborative loss leads to remunerations for the data sellers based on the contribution of their features to improve the loss, i.e., the revenue for the $j$th seller corresponds to $r_j = (\mathcal L_\text{local}- \mathcal L_\text{market})\times b_i \times \sum_{k\in \mathcal{X}_{j,t}} \psi_{j,k},$
where $\psi_{j,k}$ measures the relative contribution of the $k$th variable from the $j$th seller to improve the loss, according to Shapley values.

Recently, the work in~\cite{pierre2022regression} was extended to: {a)}~integrate Bayesian optimization that treats the parameters themselves as random variables whose distribution is inferred by prior beliefs updated as new data are observed~\cite{falconer2023bayesian}, and {b)}~integrate a correlation-aware analytics market since correlated data make the revenue allocation complex since the value of overlapping information is combinatorial~\cite{falconer2023incentivizing}.

The work in~\cite{trading22} offers a different bidding strategy where data sellers make the bids. It is called a \textit{LASSO Regression analytics Market (LRM)}. In the proposed analytics-based market mechanism, illustrated in Fig.~\ref{fig:market3}, there is a data buyer $i \in \mathcal{B}$, that wants to forecast $\left\{Y_{i,t}\right\}_{t=T+1}^{T+H}$ by using the available market data $\mathcal Z_t$.
A payment threshold is introduced, $H_j^k(u_{j,k}, \beta^i_{j,k})$, which represents the payment required by an agent $j \in \mathcal{S}$, $j \neq i$, for disclosing its $k$th feature. The coefficient $\beta^i_{j,k}$ measures the relation of agent $j$'s $k$th feature with the buyer's target variable $y_i$. The seller determines the quantity $u_{j,k}$ that reflects its reservation to sell the specific feature, considering factors such as the potential loss of accuracy, loss due to an increase in competitors' profit, data collection costs, and other considerations. The forecasting approach is then based on a LASSO regression model,
\begin{equation}\label{eq:lasso-market}
    \underset{\boldsymbol\beta}{\operatorname{argmin}} \frac{1}{T}\sum_{t=1}^T (y_{i,t} {-} \sum_j \sum_{k} X_{j,t}^k \beta^i_{j,k})^2 {+} \overbrace{\sum_{j\neq i} \underbrace{\sum_k |u_{j,k} \beta^i_{j,k}|.}_\text{$j$th seller revenue}}^\text{payment of $i$th buyer}
\end{equation}
Buyers may assign a scaling factor to the mean square error (MSE) to represent their losses in monetary terms. 

Despite these approaches advancing incentives for data sharing, significant gaps persist with limited flexibility for both buyers and sellers to set prices and computational inefficiency of Shapley value-based algorithms.

\section{Proposed Analytics Market for Time Series Forecasting}
\label{sec:proposal}

The analytics market proposed in this work allows collaborative time series forecasting. It is characterized by its bid-based interaction approach, designed to incentivize active participation from data buyers and sellers. The main goal is to enhance forecasting accuracy through collective intelligence. Although this analytics market can be directly applied to other regression problems and use cases, it is formulated using the RES forecasting notation described in Section~\ref{notation}.

\subsection{Market Entities}
\label{sec:data-market-entities}

The proposed algorithm considers three entities: data buyers, data sellers, and the market operator. A data seller is an agent $j \in \mathcal{S} \subseteq \mathcal{A}$ aiming to generate monetary revenue by submitting data $\mathcal{X}_{j,t}$ to the market. These sellers are motivated by revenue generation and lack specific knowledge about the forecasting methods that will use their data.
On the other side, a data buyer is an agent $i \in \mathcal{B} \subseteq \mathcal{A}$ with a unique regression task  $\left\{Y_{i,t'}\right\}_{t'=t+1}^{t+H}$. Without the collaborative framework, a buyer \( i \) would be constrained to approximate \( Y_{i,t} \) using a local function \( f_{\text{local}}^{i}(\boldsymbol{X}_i, \boldsymbol{\Theta}^\text{local}_i) \) built on local historical data \( \left\{\mathcal{X}_{i,t}\right\}_{t=1}^T \). Here, \( \boldsymbol{X}_i \) corresponds to the local feature matrix, and \( \boldsymbol{\Theta}^\text{local}_i \) represents the parameters to be estimated. However, with the proposed market model, the buyer \( i \) enters the market intending to enhance forecasting accuracy through a market model \( f_{\text{market}}^{i}(\boldsymbol{Z}, \boldsymbol{\Theta}^\text{market}_i) \), where \( \boldsymbol{Z} \) is the feature matrix with all historical data collected by the market \( \left\{\mathcal{Z}_t\right\}_{t=1}^T \).

At the center of this interaction lies the market operator, who orchestrates various operations. This includes acquiring data from sellers, managing an auction mechanism incorporating both buyers and sellers, performing regression tasks, computing prices, distributing the revenue among sellers, and overseeing other tasks to ensure the smooth functioning of the market. Sellers $j \in \mathcal{S}$ only possess access to their data $\mathcal{X}_{j,t}$, while buyers $i \in \mathcal{B}$ can only access forecasts generated for their time series, $Y_{i,t}$. This setup ensures data privacy, contingent upon the market operator's integrity and impartiality as a trusted intermediary. The collaborative forecasting model \(f_{\text{market}}^{i}\) is always an additive model linear in its parameters. Conversely, \(f_{\text{local}}^{i}\) can include a wide range of statistical models chosen at the discretion of the buyer \(i\).

\subsection{Gain Function and Estimation}
The gain function $\mathcal{G}_i$ relates to the improvement in forecasting accuracy achieved by using data from the market, and it is typically related to a loss function $\mathcal{L}_{i}(\cdot)$, e.g., root mean squared error. 
When a buyer $i \in \mathcal{B}$ enters the market at a timestamp $t=T$, the gain that a buyer can achieve in forecasting  $\left\{Y_{i,t'}\right\}_{t'=t+1}^{t+H}$ is not directly attainable due to the uncertainty of future outcomes. To tackle this, the historical data is divided into a training and a validation set, and the gain of agent $i$ at a time $t$ can be defined, e.g., as the percentage of loss improvement:
\begin{equation}
\label{eq:gain-definition}
    \mathcal{G}(\boldsymbol \Lambda_{i}) = \frac{\left( \mathcal{L}_{i}(\boldsymbol{X}_i, \boldsymbol{y}_i, f_{\text{local}}^i) {-} \mathcal{L}_i(\boldsymbol{Z}, \boldsymbol{y}_i, f_{\text{market}}^{i}) \right)_+}{\mathcal{L}_{i}\left(\boldsymbol{X}_i, \boldsymbol{y}_i, f_{\text{local}}^{i}\right)}{\times} 100,
\end{equation}
where the training-validation division is implicit, $(x)_+ = \max(x,0)$ and $\boldsymbol \Lambda_{i} = \{{\mathcal L_i,}\boldsymbol{X}_i, \boldsymbol{Z}, \boldsymbol{y}_i, f_{\text{local}}^{i}, f_{\text{market}}^{i}\}$. 
To manage real-time series data dynamics, future gains should be estimated by leveraging historical timestamps with similar covariates, e.g., selected through Euclidean distance.

\subsection{Bids and Price Definition}
\label{auctionmechanism}
The proposed analytics market introduces a distinctive bid-based interaction framework to foster buyer and seller engagement within the marketplace. We will investigate how the market entities are involved in this auction-based mechanism.

\subsubsection{Buyers}
The works in~\cite{carla2021} and~\cite{pierre2022regression} allow a buyer \(i \in \mathcal{B}\) to propose a public bid, \(b_i\), representing the maximum price he is willing to pay for a unit increase in the gain function \(\mathcal{G}(\cdot)\). It means a willingness to pay a maximum amount of $\mathcal{G}(\boldsymbol{X}_i, \boldsymbol{Z}, \boldsymbol{y}_i, f_{\text{local}}^{i}, f_{\text{market}}^{i}, b_i) \times b_i$.
However, a buyer valuation of gain may not be so straightforward. For instance, a buyer may assign higher values to the initial gains, signaling their eagerness to pay more for substantial improvements in their forecasting accuracy. As the gain increases, the buyer's valuation of further improvements might diminish, reflecting a diminishing marginal utility of gain. 
The proposed framework introduces a more flexible way for buyers to express their preferences and willingness to pay for improvements in forecasting accuracy. Each buyer now offers a value function $\mathcal{VF}_i(g)$ (from decision theory~\cite{Chankong2008}) that quantifies the bids as a function of the gain achieved through the analytics market.

\subsubsection{Sellers}
Agents are afforded the ability to propose sets of bids \(\boldsymbol{s}_j = (s_{j,1}, s_{j,2},\allowbreak ...,\allowbreak s_{j,n_j})\), \(j \in \mathcal{S}\), for the use of each of their variables. Each \(s_{j,k}\) denotes the monetary compensation sellers require for sharing their $k$th variable within \(\mathcal{X}_{j,t}\). 

\subsubsection{Market Operator}
\begin{figure}
    \centering
    \includegraphics[width=0.16\textwidth]{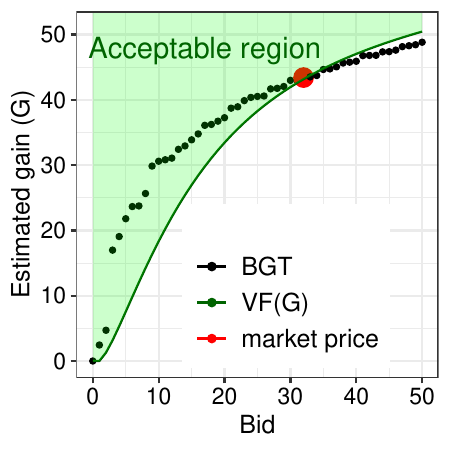} \hfill
    \includegraphics[width=0.16\textwidth]{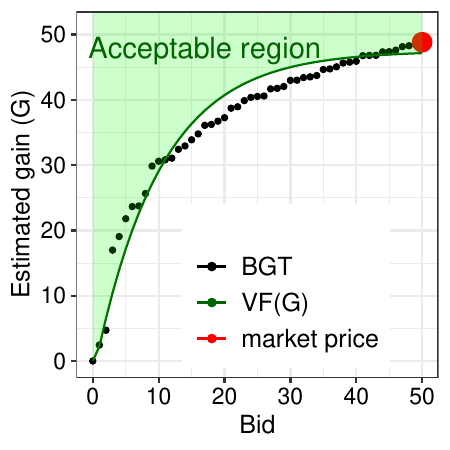}\hfill
    \includegraphics[width=0.16\textwidth]{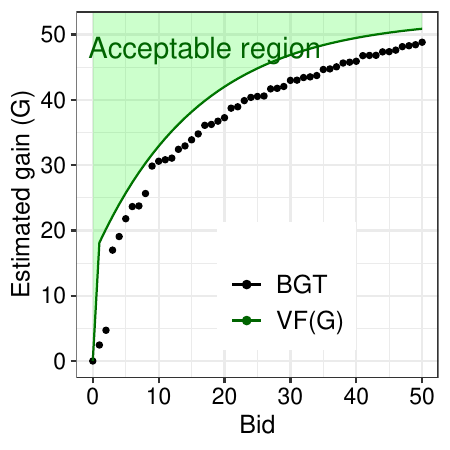}
    \caption{Illustration of price definition.}
    \label{fig:price-definition}
\end{figure}

The market operator manages the auction mechanism and determines the prices buyers should pay (denoted as $p_i$ for buyer $i$). To bridge the gap between what the market offers and the value perceived by the buyers, the market operator constructs a Bid-Gain Table (BGT) that summarizes the gain associated with each bid $b$ within a set of possible bids $\Tilde{\boldsymbol{b}} = \left[b_{min}, b_{min} + \delta_b, b_{min} + 2\delta_b, \ldots, b_{max} \right]$. This BGT is a critical component in determining the final payment,
\begin{equation}
\label{eq:final-price}
    p_i = \min_b \left\{ \underset{b}{\operatorname{argmax}} \, \mathcal{G}(\boldsymbol {\Lambda}_i, b)\right\} \, \text{s.t.} \,\,  b \leq \mathcal{VF}(\mathcal{G}(\boldsymbol {\Lambda}_i, b)),
\end{equation}
which corresponds to the bid that maximizes the expected gain while respecting the buyer valuation function. The value function represents the minimum acceptable gain for each possible price. Fig.~\ref{fig:price-definition} illustrates the price definition, where green regions are acceptable for the buyer. In the first plot, for bids up to 31, the gains estimated by the market operator are higher than the ones requested by the buyer, meaning the buyer is comfortable with such prices. For bids higher than 31, the estimated gain is lower than the one requested by the buyer, meaning the buyer is unwilling to pay a final price higher than 31. In the second plot, bids up to 10 and higher than 40 respect the minimum estimated gain required by the buyer, while the bids between 10 and 40 do not. Finally, in the third plot, none of the bids have an estimated gain that satisfies the minimum acceptable gain for the buyer. In such cases, the market operator decides on a zero price and communicates that no forecasts respect the value function.
The market price is related to the buyer's benefit, meaning the buyer only pays if there is an expected improvement in forecasting accuracy.

\subsection{Spline Bid-Constrained LASSO Regression}
\label{sec:spline-bid-constrained-lasso-regression}
The forecasting problem tackled within the proposed market must account for the restrictions imposed by the buyers' and sellers' bids and the complex relationships between the data. 
Our proposal employs a B-spline LASSO regressor aligned with the bid strategy described in Section~\ref{auctionmechanism}. After the B-spline transformer, each variable $Z \in \mathcal{Z}_t$ is transformed into a group of $M$ variables, where $M=D+K+1$ is the sum of the spline order $D$ with the number of knots $K$ plus 1, i.e., the original set
\begin{equation}
\mathcal{Z}_{t}=\{{\ldots},\underbrace{X_{i, t}^1, {\ldots}, X_{i, t}^{n_i}, Y_{i,t_0{-}1}, {\ldots}, Y_{i,t_0{-}L}}_{\text{power plant i}}, {\ldots}\},
\end{equation}
is transformed into
\begin{equation}
    \Tilde{\mathcal{Z}_t} = \{\ldots, \underbrace{X_{i,1,t}^1,\allowbreak \dots,\allowbreak X_{i,M,t}^1}_\text{group related to $X_{i,t}^1$}, \dots, \underbrace{X_{i,1,t}^{n_i},\allowbreak \dots,\allowbreak X_{i,M,t}^{n_i}}_\text{group related to $X_{i,t}^{n_i}$}, \dots\}
\end{equation}
that consists of $p$ groups of variables, where $p$ is the number of variables in $\mathcal{Z}_t$.  To align the forecasting approach with the market participants' bids, we consider seller $j$ wants to receive $s_{j,k}$ if its $k$th variable is used. Since the market applies B-splines, if at least one of the variables in the group related to the $k$th variable is used, then seller $j$ receives $s_{j,k}$. 
Therefore, for a buyer bid $b_i$, the spline LASSO regressor estimated by the market operator assumes the form
\begin{subequations}
\label{eq:argmin_bid_constrained}
\begin{equation}
\hat{\boldsymbol{\Theta}}_i = \underset{\boldsymbol\Theta_i=(\boldsymbol{\beta}^i, \boldsymbol{\eta}^i)}{\operatorname{argmin}} \, \mathcal L(\mathbf y_i, \sum \boldsymbol\Theta_i \tilde{\mathbf Z}_{i}) + \lambda\|\boldsymbol \Theta_i\|_1 \\ 
\end{equation}
\begin{equation}
\label{eq:bid-constraint}
\begin{aligned}
\text{s.t. } & \sum_{j=1}^N r_j \leq b_i,
\end{aligned}
\end{equation}
\end{subequations}
where 
\begin{equation}
    \|\boldsymbol \Theta_i\|_1 = \sum_{m=1}^M\left( \sum_{j=1}^N \left[\sum_{k=1}^{n_j}  |\beta_{j,k,m}^i| + \sum_{l=1}^L |\eta^i_{j,l,m}| \right]\right),
\end{equation}
the revenue for $j$th seller, $r_j$, is
\begin{equation}
\label{eq:revenue-sellers-proposal}
\begin{aligned}
    r_j= & \sum_{k=1}^{n_j} s_{j,k} \left[1{-}\prod_{m=1}^{M}\left(1{-}\mathcal{I}\left(\beta_{j,k,m}^i\right)\right)\right] {+} \\
      & \sum_{l=1}^L s_{j,l+n_j} \left[1{-}\prod_{m=1}^M\left(1{-}\mathcal{I}\left(\eta_{j,l,m}^i\right)\right)\right], 
\end{aligned}
\end{equation}
$\beta_{j,k,m}^i$ is the coefficient associated to the $m$th spline of the $k$th variable of agent $j$ when forecasting $Y_{i,t}$, $\eta_{j,\ell,m}^i$ is the coefficient associated with the $m$th spline of the $\ell$ most recent power value from agent $j$ when forecasting $Y_{i,t}$, and $\mathcal{I}(x){=}\{1 \text{ if } x{\neq} 0, 0 \text{ otherwise}\}$.

As proven in \cite{yu2022high}, the solution of~\eqref{eq:argmin_bid_constrained} can be obtained by iteratively solving 
\begin{equation}
\label{eq:min-theta}
\begin{aligned}
\underset{\boldsymbol \Theta_i}{\operatorname{argmin}} & \, \frac{1}{2}\|\boldsymbol{\Theta}_i {-} \boldsymbol{a}\|_2^2 {+} \lambda \|\boldsymbol \Theta\|_1 \text{ subject to }~\eqref{eq:bid-constraint},
\end{aligned}
\end{equation}
where at iteration $k$, $\boldsymbol{a}^k {=} \boldsymbol\Theta^{(k-1)} {-}\frac{1}{C} \frac{\partial \mathcal L\left(\boldsymbol\Theta\right)}{\partial \boldsymbol\Theta}(\boldsymbol \Theta^{(k-1)})$. For $\mathcal L(\cdot)=\frac{1}{2T}\sum_t \left(\mathbf y_t - \mathbf Z_t \boldsymbol\Theta_i\right)^2$, $\boldsymbol{a}^{k}{=}\allowbreak\boldsymbol \Theta^{(k-1)} {-}\allowbreak \frac{1}{T C} \boldsymbol{\Tilde{Z}}^\top \left(\boldsymbol{y}_i {+} \boldsymbol{\Tilde{Z}}\boldsymbol{\Theta}_i^{(k-1)}\right)$. 
\begin{proposition}
If $\hat{\boldsymbol{\Theta}}_i$ is an optimal solution to~\eqref{eq:min-theta}, then $$\hat{\boldsymbol{\Theta}}_i=\text{sign}(\boldsymbol a - \lambda) \circ \left(|\boldsymbol a|-\lambda\right)_{+} \circ \hat{\boldsymbol{w}},$$
where $\hat{\boldsymbol{w}}=\left(\hat{w}_{1,1} \mathbf{1}_{M}, \hat{w}_{1,2} \mathbf{1}_{M}, \ldots, \hat{w}_{j,k} \mathbf{1}_{M}, \ldots\right)^\top$, $\mathbf{1}_{M}$ is the row vector of M 1's, and $\hat{w}_{1,1}, \hat{w}_{1,2}, \ldots, \hat{w}_{j,k}$ is the solution to the following 0-1 knapsack problem:
\begin{equation}
    \underset{w_{j,k}\in\{0,1\}}{\operatorname{argmax}} \sum_{j=1}^N \sum_{k=1}^{n_j+L}\mu_{j,k} w_{j,k}  \,\,\operatorname{ s.t. } \sum_{j=1}^N \sum_{k=1}^{n_j+L}w_{j,k} s_{j,k} {\leq} b_i,
\end{equation}
where 
\begin{equation}
    \mu_{j,k} {=} \sum_{m=1}^M \frac{a_{j,k,m}^2 {-} 2\lambda |a_{j,k,m}| {+} \lambda^2}{2} \cdot \frac{1+\text{sign}(|a_{j,k,m}| {-} \lambda)}{2}.
\end{equation}
\end{proposition}
\begin{proof}
    See~\cite{yu2022high}.
\end{proof}
Algorithm~\ref{alg:knapsack} solves a knapsack problem with weights (bids from sellers) $\boldsymbol{s} \in \mathbb{R}^{p\times1}$, capacity (bid from buyer) $b_i$ and item values $\boldsymbol{\mu} \in \mathbb{R}^{p\times1}$, where $p=\sum_{j=1}^N n_j + N L$. Based on~\cite{yu2022high}, Algorithm~\ref{alg:BCR} is applied to estimate the parameters $\boldsymbol \Theta_i$. The inputs are the transformed historical data $\{\boldsymbol{y}_i, {\boldsymbol{\Tilde{Z}}}\}$, the sellers bids $\boldsymbol{s}$, the buyer bid $b_i$, a tolerance parameter $\varepsilon$ and the LASSO term $\lambda$. The spline LASSO hyperparameters, $\lambda, D$, and $K$, are estimated by grid search and per buyer bid $b_i$, as shown in Algorithm~\ref{alg:tuningBCR}. $\mathcal{ST}(\cdot)$ is a B-spline transformer and  $\mathcal{FS}(\cdot)$ is a feature selection filter based on partial Pearson correlation and corresponding hypothesis test.

\begin{algorithm}
\caption{Dynamic Programming 0-1 Knapsack $\mathcal{KS}$}
\label{alg:knapsack}
\begin{algorithmic}[1]
    \State \textbf{Input:} $\boldsymbol{s}$, $\boldsymbol{\mu}$, $b_i$
    \State \textbf{Output:} Allocation vector $I$
    
    \State $D_{[0, w]} \leftarrow 0, \forall w \in \{0, \dots, b_i\}$
    
    \For {$j = 1$ to len($\boldsymbol s$)}
        \For {$w = 0$ to $b_i$}
            \If{$s_{[j]} > w$}
                \State $D_{[j,w]} \leftarrow D_{[j-1,w]}$
            \Else
                \State $D_{[j,w]} \leftarrow \max(D_{[j-1,w]}, D_{[j-1,w - s[j]]} + \mu_{[j]})$
            \EndIf
        \EndFor
    \EndFor
    \State $ K {\gets} D_{[length(\boldsymbol s),b_i]}$, $w {\gets} p_i$, $I_{[j]} {\gets} 0\, \forall j \in \{1, {\dots}, len(\mathbf s)\}$ 
    \For{$j = \text{len}(\boldsymbol s)$ to $1$}
        \If{$K \leq 0$}
            \State \textbf{break}
        \ElsIf{$K \neq D_{[j - 1,w]}$}
            \State $I_{[j]} \gets 1$, $K \gets K - \mu_{[j]}$, $w \gets w - s_{[j]}$
        \EndIf
    \EndFor
    \State \textbf{Return} $I$
\end{algorithmic}
\end{algorithm}

\begin{algorithm}
\caption{Spline Bid-Constrained LASSO Regression}
\label{alg:BCR}
\begin{algorithmic}[1]
    \State \textbf{Input:} $\boldsymbol{y}_i$, $\boldsymbol{\Tilde{Z}}$, $\boldsymbol{s}$, $b_i$, $\varepsilon$, $\lambda$, $maxiter$
    \State \textbf{Output:} $\hat{\boldsymbol{\Theta}}_i \gets \boldsymbol{\Theta}^{(k-1)}$
    \State $\boldsymbol{\Theta}_i^{(0)} \leftarrow (\boldsymbol{\beta}^i, \boldsymbol{\eta}^i)$ such that
    \eqref{eq:bid-constraint} holds
    \State $\Psi \leftarrow \left(\boldsymbol{\Tilde{Z}}, \boldsymbol{y}_i\right)$
    \State $C \gets \lambda_{max}(\frac{1}{T} \boldsymbol{\Tilde{Z}}^\top \boldsymbol{\Tilde{Z}}) + 0.1$ {\color{gray} $\triangleright \lambda_{max}$  largest eigenvalue}
    \State $k \gets 1$
    \While{$\mathcal{L}( \Psi, \boldsymbol{\Theta}_i^{(k)}) {-} \mathcal{L}(\Psi,\boldsymbol{\Theta}_i^{(k-1)}) {>} \varepsilon$ and $k{<}maxiter$}
        \State $\boldsymbol{a}^{(k)} {\gets} \boldsymbol\Theta^{(k{-}1)} {-}\frac{1}{C} \frac{\partial \mathcal L\left(\Psi, \boldsymbol\Theta\right)}{\partial \boldsymbol\Theta}(\Psi, \boldsymbol \Theta^{(k{-}1)})$ 
        \State $\boldsymbol{\mu}_{j,k}^{(k)} {\gets} \sum_{m} \frac{({a^{(k)}_{j,k,m}})^2 {-} 2\lambda |a^{(k)}_{j,k,m}| {+} \lambda^2}{2} {\cdot} \frac{1+\text{sign}(|a^{(k)}_{j,k,m}| {-} \lambda)}{2}$
        \State $\mathbf{w}^{(k)} {\gets} \mathcal{KS}(\mathbf{s}, \boldsymbol{\mu}^{(k)}, b_i)$ \quad {\color{gray} $\triangleright$  Algorithm~\ref{alg:knapsack}}
        \State $\boldsymbol{\Theta}^{(k)} {\gets} \text{sign}(\boldsymbol a^{(k)} {-} \lambda) {\circ} \left(|\boldsymbol a^{(k)}|{-}\lambda\right)_{+} {\circ} \left(\hat{w}_{1,1} \mathbf{1}_{M}, \ldots \right)$
        \State $k \gets k+1$
    \EndWhile
\end{algorithmic}
\end{algorithm}

\begin{algorithm}
\caption{Tuning Bid-Constrained Model Hyperparameters}
\label{alg:tuningBCR}
\begin{algorithmic}[1]
    \State \textbf{Input:} $\boldsymbol{y}_i$, $\boldsymbol{Z}$, $\boldsymbol{s}$, $b_i$, $\varepsilon$, $\boldsymbol{\Omega_i}=\{\boldsymbol{D}, \boldsymbol{K}, \boldsymbol{\Lambda}\}$, $maxiter$ , $\alpha$
    \State \textbf{Output:} $D_{b_i}, K_{b_i}, \lambda_{b_i} \gets \arg_{D,K,\lambda} \min L_{D,K,\lambda}$
    \State $L_{D, K, \lambda} \gets \mathbf 0$
    \For{$D \in \boldsymbol{D}$}
    \For{$K \in \boldsymbol{K}$}
    \For{$\boldsymbol{Z}^{tr}, \boldsymbol{y}_i^{tr}, \boldsymbol{Z}^{val}, \boldsymbol{y}_i^{val}, \in$ cv\_indexes}
        \State $\boldsymbol{\Tilde{Z}}^{tr} \leftarrow \mathcal{ST} (\boldsymbol{Z}^{tr}, D, K)$ {\color{gray} $\triangleright$  transformer}
        \State $\boldsymbol{\Tilde{Z}}^{tr} \leftarrow \mathcal{FS}(\boldsymbol{\Tilde{Z}}^{tr}, \alpha)$ {\color{gray} $\triangleright$  feature selection} 
        \For{$\lambda \in \boldsymbol{\Lambda}$}
            \State $\hat{\boldsymbol{\Theta}}_i \gets$ Alg.2($\boldsymbol{y}_i^{tr}$, $\boldsymbol{\Tilde{Z}}^{tr}$, $\boldsymbol{s}$, $b_i$, $\varepsilon$, $\lambda$, $maxiter$)
            \State $L_{D, K, \lambda} \gets L_{D, K, \lambda} + \mathcal L(\Tilde{\boldsymbol{Z}}^{val},\boldsymbol{y}_i^{val},\hat{\boldsymbol{\Theta}}_i)$
        \EndFor
        
    \EndFor
    \EndFor
    \EndFor
\end{algorithmic}
\end{algorithm}

\subsection{Data-sharing Incentive Mechanism}
\label{sec:data-market-mechanism}

The proposed data-sharing incentive algorithm is depicted in Algorithm~\ref{alg:proposal} and it is called a \textit{Spline LASSO bid-Constrained Market (SLCM)}. The process begins at time \(T\) when the market operator opens a session. During this phase, sellers \( j \in \mathcal{S} \) submit their data \( \left\{\mathcal{X}_{j,t}\right\}_{t=1}^{T+H} \) and bid vectors \( \boldsymbol{s}_j = (s_{j,1}, \ldots, s_{j,n_j+L}) \) to the market operator. Simultaneously, buyers \( i \in \mathcal{B} \) provide their targets \( \left\{Y_{i,t}\right\}_{t=T}^{T+H} \) and value functions $\mathcal{VF}_i$.

{
\begin{algorithm}
\caption{Proposed mechanism at time $t=T$.}
\label{alg:proposal}
\begin{algorithmic}[1]
\For {each market session}
        \State Define forecasting horizon $h=1,\dots, H$
	\State Market session state = 'open'
        \While {Market session state = 'open'}
		\For {each buyer $i$}
			\State Submit data $\{Y_{i,t}\}_{t=1}^T, \{\mathcal{X}_{i,t}\}_{t=1}^{T+H}$
			\State {\color{red}Submit bid function $\mathcal{VF}_i(g)$}
		\EndFor
		\For {each seller $j$}
		 \State Submit data $\{\mathcal{X}_{j,t}\}_{t=1}^{T+H}$
		 \State {\color{red}Submit bids per feature $\boldsymbol s_j$} 
		\EndFor
	\State Update Market session state = 'closed'
	\EndWhile
	\While{Market session state = 'closed'}
		\For {each buyer $i$}
			\State {\color{red}Estimate $f_{local}^i(\boldsymbol X_i, \boldsymbol \Theta_i)$ {\color{gray} $\triangleright$  Bayesian Opt.} }
			
			\If {$\{Y_{i,t}\}_{i=1}^T$ is stationary} 
				\State BGT$_i \leftarrow \{\}$ {\color{gray} $\triangleright$  unique BGT for buyer $i$} 
				\For {each bid $b$ in $\tilde{\boldsymbol b}$}
					\State {\color{red}Estimate \( {f}_{\text{market}}^{i}\) by Algorithms~\ref{alg:BCR} and \ref{alg:tuningBCR}}
					\State Compute gain $g(b)$, e.g., as in~\eqref{eq:gain-definition}
					\State Concatenate pair $(b,g(b))$ to BGT$_i$
				\EndFor
				
			\Else
				\State BGT$_{i,h} \leftarrow \{\}, h=1, \dots, H$ 
				\For {each bid $b$ in $\tilde{\boldsymbol b}$}
					\State {\color{red}Estimate \( {f}_{\text{market}}^{i}\) by Algorithms~\ref{alg:BCR} and \ref{alg:tuningBCR}}
					\For {$h = 1$ to $H$}
						\State Compute distance $d(\tilde{\boldsymbol Z}_{t+h}, \tilde{\boldsymbol Z}_{t'})$
						\State Select $k$th most similar timestamps
						\State Compute $g(b)$ for these timestamps
						\State Concatenate $(b,g(b))$ to BGT$_{i,h}$
					\EndFor
				\EndFor
			\EndIf
			\If {$\{Y_{i,t}\}_{t=1}^T$ is stationary} 
				\State Payment $p_i \gets \sum_{j} r_j$, as in~\eqref{eq:final-price}
				\State $\left\{\hat{Y}_{i,t}\right\}_{t{=}T{+}1}^{T{+}H} \gets  f_{\text{market}}^{i}( \tilde{\boldsymbol{Z}},\boldsymbol{\Theta}_i, p_i)$
			\Else
				\State $p_i, r_j \gets 0$
				\For {$h = 1$ to $H$}
					\State Payment $p_{i,h} \gets \sum_{j} r_{j,h}$ using BGT$_{i,h}$
					\State $p_i \gets p_i  + p_{i,h}, \, r_j \gets r_j + r_{j,h}$ 
					\State $\hat{Y}_{i,t+h} \gets f_{\text{market}}^{i}( \tilde{\boldsymbol{Z}}, \boldsymbol{\Theta}_i, p_{i,h})$
				\EndFor
			\EndIf

			\State Buyer pays $p_i$ and gets $\left\{\hat{Y}_{i,t}\right\}_{t{=}T{+}1}^{T{+}H}$
			\State Seller $j$ receives $r_j$, as in~\eqref{eq:revenue-sellers-proposal}
		\EndFor
	\EndWhile
\EndFor

\end{algorithmic}
{\footnotesize {\color{red}Red} instructions execute during ``re-estimation'' sessions. The dynamic pricing model adapts to time series changes but lacks real-time coefficient updates. Frequency of ``re-estimation'' sessions depends on data variability.}
\end{algorithm}
}

Following this stage, a closed session ensues, and the market operator handles all regression tasks in parallel. For each task, the following steps unfold. Firstly, the market operator performs the local model $f_{local}^{i}(\boldsymbol{X}_i, \boldsymbol{\Theta}_i)$ using the forecasting model $f_{local}^i$ chosen by the buyer $i$, trained with historical data \( \mathbf X_i \). Subsequently, for each bid \( b \) in \( \boldsymbol{\tilde{b}} \), the market spline LASSO-constrained regression model \( f_{\text{market}}^{i}\left(\boldsymbol{Z}, \boldsymbol{\Theta}_i\right) \) is computed, and its gain function values are evaluated using the loss function $\mathcal{L}(\cdot)$ given by buyer $i$, to establish the relation between forecasting skill improvement and bids. This information is presented in $BGT_i$. With this table, the market operator computes the final price to be paid by agent $i$ as in~\eqref{eq:final-price}.
The market recalls the market coefficients $\boldsymbol\Theta_i = \left\{\boldsymbol{\beta}^{i}, \boldsymbol{\eta}^{i}\right\}$ associated with the price $p_i$ to be able to compute the revenues for each seller $j$ that are given by~\eqref{eq:revenue-sellers-proposal}.
Finally, the buyer $i$ pays $p_i$ and receives the forecasts for $\left\{{Y}_{i,t}\right\}_{t=T+1}^{T+H}$.
Please note that the LASSO penalty within the data market framework facilitates feature selection, which is regulated by the importance of the features for the regression task and the sellers' bids. Sellers receive what they bid for data usage; bidding a high price value may result in unused data, yielding no revenue. Moreover, if multiple agents possess similar data, their bids dictate participation in a given forecasting task. As for buyers, the market enables them to express willingness to pay based on gain, ensuring payment aligns with their bids. Lastly, the revenue division respects the following properties:
\begin{itemize}
    \item \textit{Budget balance:} payment is total revenue, $p_i{=}\sum_j r_j.$
    \item \textit{Individual rationality}: sellers prefer market participation over non-participation, $r_j\geq 0$, $r_j=s_j$ if data are used.  
    \item \textit{Zero-element:} a data seller that provides data with zero marginal contribution to all coalitions of other features receives $r_j=0$. A data buyer with no benefit pays $0$.
    \item \textit{Truthfulness:} Data sellers receive maximum payment when sending their true data. If someone sells data with noise, $X_{j'}=X_j + \epsilon$, then the revenue $r_j \geq r_{j'}$, i.e., there is a risk of not selling data (see Appendix for a proof).
    \item \textit{Robustness to replication:} If $X_j$ and $X_{j'}$ are similar, then LASSO selects one of them, i.e.,  $r_j+r_{j'} \leq r_j$.
\end{itemize}


\section{Case Study} \label{CaseStudy}

\graphicspath{ {./Figures/CaseStudy} }

The proposed data market is evaluated on synthetic and open wind power data and compared with two state-of-the-art benchmarks where only one of the parties bids, i.e., data buyers~\cite{carla2021} or data sellers~\cite{trading22}. The hyperparameters are estimated through Bayesian optimization and grid search, and minimizing the Root MSE (RMSE)\footnote{Implementations are done in Python programming language. Source code available at \url{https://github.com/INESCTEC/budget-constrained-collaborative-forecasting-market}}.

\subsection{Basic synthetic data setup}

Two models are used to generate data: {a)} linear $y_t = \sum_{i=1}^{100} x_{i,t}\beta_i + \epsilon_t$ and {b)} non-linear $y_t = \exp\left({0.05}\sum_{i=1}^{100} x_{i,t}\beta_i\right) + \epsilon_t$. Here, $\epsilon_t$ is white noise, data buyer owns ${x_{1,t}, \dots, x_{10,t}}$, and only 10 out of the 100 covariates have $\beta_i\neq 0$ (specifically, $x_{3,t}$ and $x_{7,t}$ from the data buyer, and $x_{12,t}$, $x_{21,t}$, $x_{31,t}$, $x_{37,t}$, $x_{48,t}$, $x_{51,t}$, $x_{63,t}$, and $x_{90,t}$ from data sellers). Notably, $x_{3,t}$ and $x_{73,t}$ are redundant, as are $x_{37,t}$ and $x_{74,t}$. The variables are generated from a standard Gaussian distribution, and their relevance is shown in the results (Table~\ref{tab:synthetic-reg-aloc}).

Sellers bid at 10 cents, except for $x_{37,t}$, which costs 11 cents. For buyers, two simple value functions are tested:
\begin{enumerate}[label=\bf \#\arabic*]
    \item $\mathcal{VF}(g)=50$, i.e.,
the buyer budget is not enough to buy all relevant features (only five of eight).
    \item $\mathcal{VF}(g)=100$, i.e., the budget is enough to allocate necessary and unnecessary features.
\end{enumerate}

The relevance of features, identified by their index, for the bid-constrained model, as fitted by the market operator, is illustrated in parentheses in Table~\ref{tab:synthetic-reg-aloc}. The ``Optimal'' column orders feature indexes by their actual relevance, determined by permutation importance, as the true model is known. When faced with budget constraints, the approach prioritizes the most relevant features. In the non-linear scenario, differences between the actual relevance of features and those identified by the spline LASSO are more evident, and features with low relevance are not selected even when the budget allows it. The LASSO regularization captures redundancies, and the budget constraint facilitates the selection of the most economical options (74 is selected instead of 37).

Table~\ref{tab:synthetic-markets-comparison} compares the proposed mechanism with the two benchmarks. The objective is to evaluate the revenues of data sellers and the gain of the buyer (percentage of relative improvement in RMSE), considering two cases.
In case~\#1, the optimal collaborative model costs 80 cents, but the buyer budget is 50. This setup is only possible in ZRM~\cite{carla2021} since in LRM~\cite{trading22} there is no direct bid from the buyer. The proposal allocates the five most relevant variables, while ZRM allocates all variables, but with the addition of noise proportional to the difference between the buyer's bid and the market-established price, $noise(X_i)$ $\sim\mathcal{N}\left(0, \frac{80-50}{80}\times std(X_i)\right)$. Although ZRM provides access to more data, the distortion caused by the noise penalizes the accuracy of the collaborative model.

In case~\#2, the optimal collaborative model price is 80, and the buyer is willing to pay this amount. The goal is to evaluate how the three algorithms distribute the revenues when the buyer's gain and price are similar. The proposal allocates only the variables the budget allows to purchase, and the sellers whose data was allocated receive the bid quantity. In this case, only the relevant variables are allocated, and redundant data is excluded. Algorithms ZRM and LRM allocate more information than necessary, but the largest contributions remain correctly captured. One difference between ZRM and LRM is that in ZRM, variables 37 and 74 receive a similar reward, whereas LRM tends to select one of the variables due to the LASSO regularization. In LRM, data sellers bid the same quantities as in our proposal, forcing the algorithm to choose variable 74, whose LASSO regularization term is smaller.

\begin{table}
\centering
\caption{Feature (index) allocation and corresponding relevance (\%) for synthetic datasets.}
\label{tab:synthetic-reg-aloc}  
\addtolength{\tabcolsep}{-2pt} 
\begin{tabular}{lll|lll}
  \hline
  \multicolumn{3}{c}{\textbf{Linear}} & \multicolumn{3}{|c}{\textbf{Non-linear}} \\ 
  \textbf{Optimal} & \textbf{Case \#1} & \textbf{Case \#2} & \textbf{Optimal} & \textbf{Case \#1} &\textbf{ Case \#2}  \\ 
  \hline
21 (24.07) & 21 (28.39) & 21 (24.06) & 21 (27.14) & 21 (32.89) & 21 (31.85)\\ 
90 (21.62) & 90 (25.03) & 90 (21.62) & 90 (23.13) & 90 (24.87) & 90 (23.89)\\ 
63 (18.68) & 63 (20.96) & 63 (18.65) & 63 (20.3) & 63 (21.05) & 63 (19.93)\\ 
48 (16.1) & 48 (17.53) & 48 (16.14) & 48 (15.44) & 48 (15.76) & 48 (15.16)\\ 
37 (8.53) & 74 (8.09)  & 74 (8.51) & 37 (7.14)   & 74 (5.43) & 74 (4.83)\\ 
51 (6.06) & --         & 51 (6.05) & 51 (4.73)   & -- & 51 (3.35)\\ 
12 (3.59) & --         & 12 (3.58) & 12 (1.68)   & -- & 12 (1.00)\\ 
31 (1.37) & --         & 31 (1.38) & 31 (0.44)   & -- & -- \\
others (0)  & -- & -- & others (0) & -- & --\\
\hline
\end{tabular}
\medskip
\centering
\caption{Comparison with benchmarks, regarding sellers revenue and buyer gain, for synthetic datasets.}
\label{tab:synthetic-markets-comparison}  
\addtolength{\tabcolsep}{-3.2pt} 
\begin{tabular}{l|rr|rr|rrr|rrr}
  \hline
  & \multicolumn{4}{c}{\textbf{Case \#1}} & \multicolumn{6}{|c}{\textbf{Case \#2}} \\ 
  &\multicolumn{2}{c|}{\textbf{Linear}} & \multicolumn{2}{c|}{\textbf{Non-linear}} &
    \multicolumn{3}{c|}{\textbf{Linear}} & \multicolumn{3}{c}{\textbf{Non-linear}}\\
    \hline
  $i$ & SLCM & ZRM  & SLCM & ZRM & SLCM & ZRM & LRM& SLCM & ZRM & LRM \\ 
  \hline
21 &  10.00 & 9.60 &  10.00 & 2.77 & 10.00 & 19.50 & 17.81 & 10.00 & 24.09 & 17.09 \\ 
90 &  10.00 & 11.80 &  10.00 & 2.47 & 10.00 & 18.14 & 16.12 & 10.00 & 15.64 & 14.49 \\ 
63 &  10.00 & 8.93 &  10.00 & 1.53 & 10.00 & 16.13 & 14.08 & 10.00 & 14.44 & 13.12 \\ 
48 &  10.00 & 6.75 &  10.00 & 2.44 & 10.00 & 13.48 & 12.31 & 10.00 & 15.69 & 11.40 \\ 
74 &  10.00 & 1.01 &  10.00 & 0.40 & 10.00 & 1.94 & 7.34 & 10.00 & 1.82 & 6.24 \\ 
51 & 0.00 & 0.99 & 0.00 & 0.02 & 10.00 & 2.48 & 5.28 & 0.00 & 1.33 & 5.11 \\ 
12 & 0.00 & 0.00 & 0.00 & 1.04 & 10.00 & 1.21 & 3.51 & 0.00 & 1.94 & 2.49 \\ 
31 & 0.00 & 0.38 & 0.00 & 0.01 & 10.00 & 0.42 & 1.82 & 0.00 & 0.00 & 1.43 \\ 
37 & 0.00 & 0.85 & 0.00 & 0.36 & 0.00 & 1.83 & 0.33 & 0.00 & 1.63 & 0.00 \\ 
others & 0.00 & $<$0.99 & 0.00 & $<$3.00 & 0.00 & $<$0.46 & $<$0.09 & 0.00 & $<$0.32 & $<$0.42 \\ 
    \hline
Gain & 78.01 & 61.43 & 25.79 & 18.64 & 93.96 & 93.01 & 93.23 & 26.71 & 26.12 & 26.43 \\
 \hline
\end{tabular}
\end{table}

\subsection{Advanced synthetic data setup}

Multiple datasets are simulated with varying number of features (100, 200, 300, 400, 500), model sparsity (25\%, 50\%), and covariance matrix sparsity (0\%, 25\%) to evaluate the robustness and scalability of our proposal. Ten simulations per combination are performed, and the impact of feature selection is also analyzed.

\begin{figure}
    \centering
    \includegraphics[width=\linewidth]{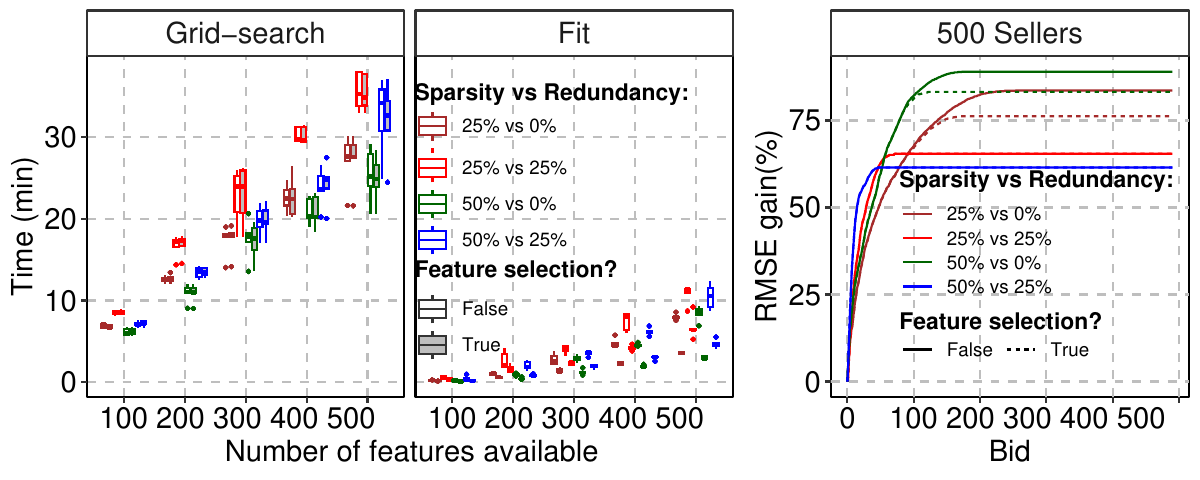}
    \caption{Results for the advanced synthetic setup.}
\label{fig:advanced_results_time}
\end{figure}

Fig.~\ref{fig:advanced_results_time} shows computational times, with hyperparameter estimation being the most time-intensive, followed by budget-constrained model training. Constructing BGT and defining prices is almost instantaneous. Filter selection reduces computational times, and increased sparsity amplifies feature selection's effect. Fig.~\ref{fig:advanced_results_time} also shows the RMSE improvement for 1 repetition, using 500 data sellers and a linear relationship. With a significance level of 5\%, the filter selection method maintains accuracy while reducing computation time. Accuracy improves with higher bids, stabilizing after a certain threshold.

\subsection{Wind power data}

The goal is to produce 24-hour-ahead wind power generation forecasts for 10 Australian zones (or wind power plants), with a launch time 00h00. The dataset used in the Global Energy Forecasting Competition 2014 (GEFCom2014)~\cite{Hong2016} spans Jan 1, 2012, to Nov 30, 2013. Hourly wind power measurements are normalized by zone capacities. Predictors include 24-hour-ahead forecasts for zonal and meridional wind components at 10m and 100m above ground level, issued daily at 00h00 to production locations.
Each zone acts as a \textit{data prosumer}, i.e., an agent that consumes and supplies data to the market ($\mathcal{A} = \mathcal{B} = \mathcal{S}=\{1,\ldots,10\}$). Agent $i \in \mathcal{A}$ owns four exogenous variables $X_{i,t}^1, X_{i,t}^2, X_{i,t}^3, X_{i,t}^4$ corresponding to the zonal and meridional wind components at 10m and 100m.

\subsubsection{Data analysis and feature engineering}

\begin{figure}
\centering
\includegraphics[trim={0.25cm 1.3cm 0.45cm 0.45cm}]{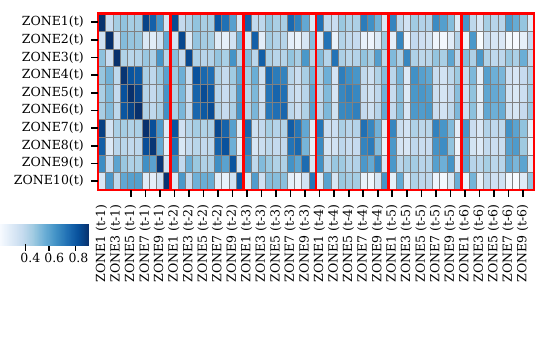}
\caption{Cross-correlation between zones.}
\label{fig:pccfs-gefcom2014}
\end{figure}

Fig.~\ref{fig:pccfs-gefcom2014} shows the cross-correlations between zones.
As lags increase, correlation decreases, revealing clusters: Zones 4, 5, and 6 are strongly correlated, as are Zones 7 and 8. Based on this analysis, lags up to 6h were considered. Table~\ref{tab:laggedfeatures} summarizes the features used per forecasting horizon; seven models are used to align with the power values available at 00h00.

\begin{table}
\setlength{\tabcolsep}{3.5pt}
\caption{Features for forecasts generated at day D+1 00:00.}
\label{tab:laggedfeatures}
\centering
\begin{tabular}{l|l|lllll|l}
\hline
Day D+1 & $D+1$ & \multicolumn{5}{c|}{Day $D$} & Exog.\\
Target (t) & 00:00 & 23:00 & 22:00 & 21:00 & 20:00 & 19:00 & X(t) \\
\hline
01:00 & \checkmark & \checkmark & \checkmark & \checkmark & \checkmark & \checkmark & \checkmark \\
02:00 & \checkmark & \checkmark & \checkmark & \checkmark & \checkmark & -- & \checkmark\\
03:00 & \checkmark & \checkmark & \checkmark & \checkmark & -- & -- & \checkmark\\
04:00 & \checkmark & \checkmark & \checkmark & -- & -- & -- & \checkmark\\
05:00 & \checkmark & \checkmark & -- & -- & -- & -- & \checkmark\\
06:00 & \checkmark & -- & -- & -- & -- & -- & \checkmark\\
$>$06:00 & -- & -- & -- & -- & -- & -- & \checkmark\\
\hline
\end{tabular}
\end{table}
  
\subsubsection{Comparison of forecasting models in Table~\ref{tab:models-evaluated}}
\label{sec:comp-models}

\begin{table}
\centering
\caption{Evaluated models and tuned hyperparameters.}
\label{tab:models-evaluated}
\begin{tabular}{p{0.95cm}p{5.05cm}p{1.5cm}}
    \hline
    \textbf{Model} & \textbf{Hyper-parameters} & \textbf{Range} \\
    \hline
    LASSO & Regularization ($\lambda$) & $[10^{-3},100]$ \\
    \hline
    Spline  & Regularization ($\lambda$) & $[10^{-3},100]$ \\
    LASSO   & B-splines degree ($D$) & $\left\{1, \ldots, 7\right\}$ \\
    (SLASSO) & Number of knots ($K$) & $\left\{3, \ldots, 30\right\}$ \\
    \hline
    Kernel & Regularization ($\alpha$) &  $[10^{-3},100]$ \\
    Ridge~{(KR)} & Gamma ($\gamma$) & $[10^{-3}, 1000]$ \\
    \hline
    Gradient  & Maximum depth ($max\_depth$) & $\left\{3, \ldots, 10\right\}$ \\
    Boosting & Learning rate ($\eta$) & $[10^{-3},0.5]$ \\
    Regressor & Max number of features ($max\_features$) & $\left\{1, \ldots, \sqrt{p}\right\}$ \\
    Trees & Min samples to split ($min\_samples\_split$) & $\left\{10, \ldots, 50\right\}$ \\
    (GBRT) & Min samples by leaf ($min\_samples\_leaf$) & $\left\{10, \ldots, 50\right\}$ \\
    & Samples fraction by learner ($subsample$) & $[0.7, 0.9]$ \\
    & Number of boosting stages ($n\_estimators$) & $\left\{100,\ldots,800\right\}$ \\
    \hline
\end{tabular}
\end{table}

\begin{figure}
    \centering
        \includegraphics[trim={0.1cm 0.3cm 0.1cm 0.4cm}, clip,width=\linewidth]{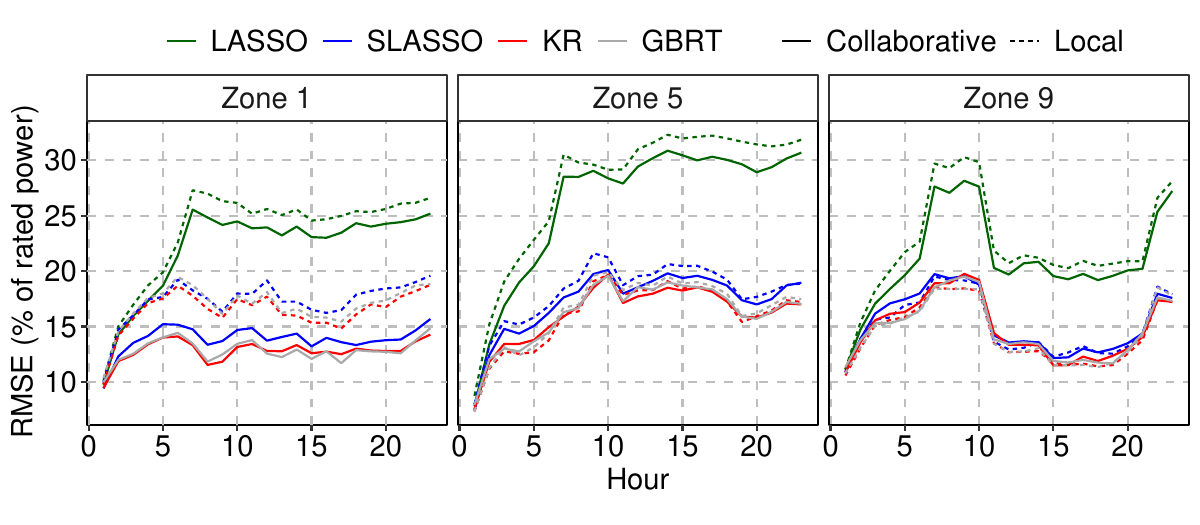}
    \caption{Comparison of forecasting models regarding RMSE.}
    \label{fig:wind-mae-rmse}
        \centering
        \includegraphics[trim={0.1cm 0.25cm 0.1cm 0.62cm}, clip,width=\linewidth]{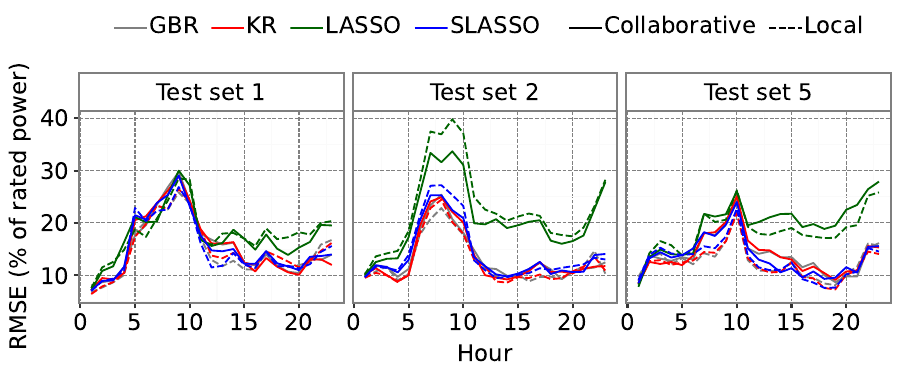}
    \caption{RMSE for three test sets, considering Zone 9.}
    \label{fig:wind-zone9}
\end{figure}

The hyperparameters are tuned using Bayesian optimization and minimizing the RMSE averaged in a 12-fold cross-validation. A sliding window of 1-month test and 12-month train is used, i.e., each model is optimized 11 times for each zone, including the hyperparameters. Fig.~\ref{fig:wind-mae-rmse} shows the RMSE per hour (for a subset of zones), averaged by the 11 test folders, where collaborative models are depicted with continuous lines, and models using only local data are depicted with dashed lines. For Zones 1 and 5, collaborative models outperform their local counterparts, but the extent of this advantage varies across zones, underscoring the potential benefits of data-sharing among zones and the need for effective incentives since the degree of such improvements varies geographically. In Zone 9, the results suggest that collaboration benefits only the linear LASSO model, which may seem unexpected. A fold-specific analysis (see Fig.~\ref{fig:wind-zone9}) reveals that in folds where collaboration is effective, the gains for LASSO are particularly pronounced, impacting the overall average in Fig.~\ref{fig:wind-mae-rmse}. 

The Gradient Boosting Regressor Trees (GBRT) and Kernel Ridge (KR) regression are front-runners among the evaluated models, consistently delivering lower RMSE. Notably, spline LASSO, an additive linear model, closely matches the performance of GBRT and KR regression, suggesting that linear models can effectively forecast RES.

Regarding average computational times, the KR exhibits the longest duration at 801 seconds for training and forecasting phases, trailed by the GBRT at 414 seconds. Conversely, the spline LASSO regression model completes forecasting tasks in 93 seconds. This highlights the practicality and feasibility of the spline LASSO regression model, which maintains competitive forecasting capabilities while minimizing computational time. These properties align with the search for accurate green artificial intelligence algorithms.

\subsubsection{Data-sharing incentive mechanism simulation}
\begin{figure}
    \begin{subfigure}{0.5\linewidth}
        \centering
        \includegraphics[width=0.93\linewidth]{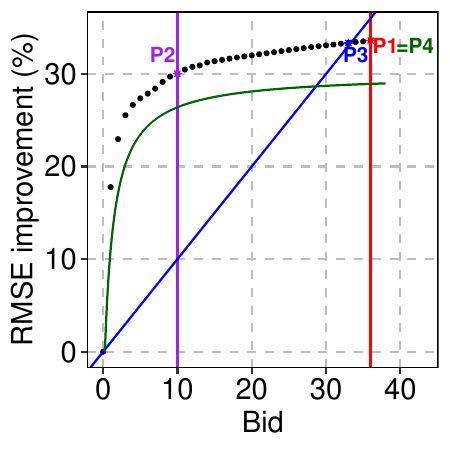}
        \caption{Illustration of price definition.}
        \label{fig:Z6S1}
    \end{subfigure}%
    \begin{subfigure}{0.5\linewidth}        \centering\includegraphics[width=0.93\linewidth]{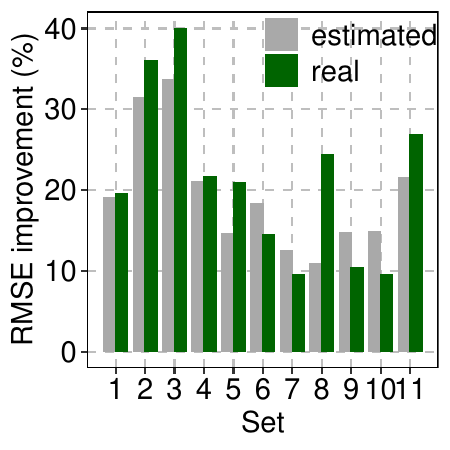}
        \caption{Estimated vs observed ($\mathcal{VF}^1)$.}
        \label{fig:Z1S4}
    \end{subfigure}
    \caption{Price definition and obtained gains for Zone 1.}
\label{fig:price-definition-case-study}
\end{figure}

A spline LASSO regression is used for local and collaborative models ($f_{local}^i$ and $f_{market}^i$), and hyperparameters are estimated by Bayesian optimization, minimizing the RMSE in a 12-fold cross-validation. While seller bids may vary, a constant bid, $s_{j,k} = 1$, is considered $\forall j,k$. Algorithm~\ref{alg:BCR} initializes $\boldsymbol{\Theta}_i^{(0)}$ by equating the coefficients of the local variables $\mathcal{X}_{i,t}$ to those of their local model $f_{local}^i$, while setting all other coefficients to zero. Four value functions are explored to define the price, depicted in Fig.~\ref{fig:Z6S1}, $\mathcal{VF}^1(g) = 100, \mathcal{VF}^2(g) = 10, \mathcal{VF}^3(g) = g$ and $\mathcal{VF}^4(g) = 40/(30-g)-1.1$. 

As wind power pattern varies over time, estimating the gain with the most recent observations can result in significant deviations between the estimated and observed gains. To address this issue, gains are estimated with the $k$ most similar historical data (after applying the spline transformer). Fig.~\ref{fig:Z1S4} shows the cumulative estimated and observed gains for Zone 1, using $k=10$.

\begin{table*}
\centering
\setlength{\tabcolsep}{1.5pt}
\caption{Cumulative payments, revenues and gain (\%) over 11 months.}
\label{tab:cumulative-payments-gains}
\begin{tabular}{r|rrrr|rrrr|rrrr|r|rr|rrr}
  \hline
   & \multicolumn{12}{c|}{\textbf{SLCM}} & \multicolumn{3}{c|}{\textbf{ZRM~\cite{carla2021}}} & \multicolumn{3}{c}{\textbf{LRM~\cite{trading22}}}\\
   \hline
   & \multicolumn{4}{c|}{Payment ($\uparrow$)} & \multicolumn{4}{c|}{Revenue ($\downarrow$)}& \multicolumn{4}{c|}{Mean gain ($\mathcal G(\%)$)} &  \multicolumn{1}{c|}{$\downarrow$\textsuperscript{*}} &  \multicolumn{1}{c}{$\uparrow$}&  \multicolumn{1}{c|}{$\downarrow$} &   \multicolumn{1}{c}{$\uparrow$} & \multicolumn{1}{c}{$\downarrow$}&   \multicolumn{1}{c}{$\mathcal G(\%)$}\\
$i$ & $\mathcal{VF}^1$ & $\mathcal{VF}^2$ & $\mathcal{VF}^3$ & $\mathcal{VF}^4$ & $\mathcal{VF}^1$ & $\mathcal{VF}^2$ & $\mathcal{VF}^3$ & $\mathcal{VF}^4$& $\mathcal{VF}^1$ & $\mathcal{VF}^2$ & $\mathcal{VF}^3$ & $\mathcal{VF}^4$ &  & \multicolumn{2}{c|}{$b_i=p_i^*=1$}  & \multicolumn{3}{c}{$u_{k,j}$=1} \\ 
  \hline

1 & 157941 & 22613 & 137907 & 60296 & 103211 & 17487 & 97387 & 44119 & 25.12 & 24.73 & 24.97 & 29.17 & 97507 & 139261 & 92582 & 61017 & 51313 & 32.43 \\ 
2 & 142742 & 29807 & 132772 & 59504 & 108121 & 22791 & 102774 & 47721 & 15.42 & 16.44 & 15.49 & 20.61 & 103888 & 118335 & 100037 & 37350 & 55567 & 10.88 \\ 
3 & 93827 & 14841 & 95458 & 33192 & 113268 & 21878 & 106349 & 48788 & -1.87 & -3.83 & -1.86 & -0.12 & 98594 & 63844 & 99296 & 49969 & 55710 & -2.00 \\ 
4 & 102468 & 22132 & 88939 & 47110 & 107611 & 10002 & 102073 & 43326 & 20.42 & 21.17 & 20.41 & 27.66 & 105260 & 109300 & 98779 & 59428 & 39475 & 15.97 \\ 
5 & 90894 & 14798 & 75095 & 44988 & 111547 & 18793 & 106193 & 46274 & 25.10 & 25.80 & 25.10 & 35.91 & 105054 & 79762 & 102226 & 55799 & 37894 & 9.22 \\ 
6 & 109366 & 18966 & 110560 & 42373 & 107990 & 18288 & 100552 & 44986 & 19.41 & 19.83 & 19.43 & 29.22 & 105731 & 83123 & 102999 & 51718 & 66962 & 8.36 \\
7 & 71848 & 12380 & 72478 & 38203 & 111038 & 21017 & 102800 & 44847 & 9.62 & 9.20 & 9.61 & 15.14 & 92765 & 81765 & 86964 & 52556 & 50787 & 6.70 \\
8 & 56730 & 10185 & 55863 & 36400 & 114681 & 28750 & 107278 & 47615 & 4.59 & 3.14 & 4.51 & 8.92 & 90387 & 71524 & 86369 & 47045 & 50435 & 3.30 \\ 
9 & 116838 & 25235 & 106607 & 48752 & 109046 & 19486 & 103044 & 46712 & 10.52 & 11.88 & 10.55 & 16.22 & 101796 & 116703 & 96514 & 39127 & 45019 & 12.98 \\
10 & 150819 & 32551 & 152278 & 52924 & 106960 & 25016 & 99507 & 46560 & 11.39 & 11.62 & 11.41 & 18.55 & 97962 & 100206 & 98058 & 47981 & 48828 & 6.62 \\ \hline
   \multicolumn{19}{l}{* Assumption: payment and gain are the same as $\mathcal{VF}^1$, whose budget allows buying all variables available on the market.}\\
\end{tabular}
\medskip
\end{table*}

Table~\ref{tab:cumulative-payments-gains} summarizes the cumulative payments, revenues, and mean gains (\%) across all zones and value functions, spanning from January 1, 2013, to November 30, 2013. Zone 1 has higher payments alongside the highest gains. Although some regions have a negative average gain, the revenue generated from information sharing compensates for this shortfall. The $\mathcal{VF}^1$ value function allows the market operator to allocate all variables in the market, meaning the buyer can obtain the best forecasts, resulting in the largest gains. Conversely, the $\mathcal{VF}^2$ function represents a constrained budget, leading to smaller gains. The value function $\mathcal{VF}^3$ represents a buyer who pays $p$ if the gain is at least $p\%$, and the results show that such a valuation allows buyers to buy the most accurate forecasts. $\mathcal{VF}^4$ represents the cases where the initial gains result in significant increases in value, but as the gains continue, the incremental increase in value decreases. The results show that this evaluation does not always allow purchasing forecasts from the best collaborative model, i.e., the gains and payments are smaller. 

To simulate the ZRM benchmark~\cite{carla2021}, initially, each zone pays the value defined for $\mathcal{VF}^1$, aiming to evaluate the revenue distribution under the scenario where buyer payment aligns with our proposal (Table~\ref{tab:cumulative-payments-gains}, 14th column). The results confirm one of the algorithm's properties~\cite{carla2021}: similar data should yield similar rewards. Zones 4, 5, and 6 demonstrate similar revenues, supported by the significant correlation between these zones (refer to Fig.~\ref{fig:pccfs-gefcom2014}). Similar conclusions are drawn for Zones 7 and 8. The second simulation performed for ZRM assumes the following: 1) buyers' bid is one, i.e., buyers should pay one monetary unit for each 1\% improvement in forecasting accuracy, and 2) the market price is also 1, i.e., no noise is added to the collaborative dataset. This configuration resembles $\mathcal{VF}^3$, and since no noise is added to collaborative data, the gains are similar to those obtained in SLCM. Compared to our proposal, buyers with higher gains pay slightly more, while those with lower gains pay less. Conversely, some sellers earn less, indicating that their features yield less than one monetary unit per timestamp.

Regarding the LRM benchmark~\cite{trading22}, a configuration similar to $\mathcal{VF}^3$ is also assumed, i.e., sellers bid 1, and the data buyer maximizes the RMSE improvement (\%). The payments and revenues are much smaller since $|\Theta_i|<10^{-3}$. Although this configuration benefits data buyers, who pay less, it may not be very convincing for data sellers. 

\section{Discussion of Potential Extensions}\label{sec:classification}

The proposed data-sharing incentive mechanism from Section~\ref{sec:proposal} can be extended to other models, namely: a) extending the bid-constrained model for cases with mixed-effects (cross-products between features), b) classification problems, and c) merging with the idea from~\cite{trading22} where sellers receive according to their coefficients.

\subsection{Mixed-effects}\label{sec:polynomial-models}

In Section~\ref{sec:spline-bid-constrained-lasso-regression}, each feature is treated with splines individually, but the approach can also incorporate mixed effects; e.g., if features $X_1$ and $X_2$ are available, with sellers' bids $r_1$ and $r_2$, the market operator can forecast a variable $Y$, with a budget $b$, by solving the budget-constrained model, 
\begin{equation} \label{eq:mixed-effects}
    \begin{aligned}
        \underset{\beta_1,\beta_2, \beta_3}{\operatorname{argmin}} & \,\frac{1}{T}\sum_t(Y_t- X_{1,t}^2\beta_1 -  X_{2,t}^2\beta_2 - X_{1,t} X_{2,t}\beta_3)^2   \\
        \text { s.t. }  & s_1 \left[1-(1-\mathcal{I}\left(\beta_1\right))(1- \mathcal{I}\left(\beta_3\right)) \right] {+}\\ 
        & s_2 \left[1-( 1-\mathcal{I}\left(\beta_2\right))(1- \mathcal{I}\left(\beta_3\right))\right] {\leq} b.
    \end{aligned}
\end{equation}
The problem~\eqref{eq:mixed-effects} can be solved using the reasoning in Section~\ref{sec:spline-bid-constrained-lasso-regression}, where price allocation relates to a group of coefficients. As shown in~\cite{gauthier2021next}, a linear combination of mixed effects among (shifted) time series can serve as a powerful model within reservoir computing, effectively capturing dynamic system data through observed time series.

\subsection{Classification Problems}\label{sec:classification-method}

For the algorithm in Section~\ref{sec:proposal}, only the loss function and forecasting model need adjustments for logistic regression. Classification markets can be used, e.g., for wind turbine fault detection using spatio-temporal data~\cite{qian2023wind}.
\subsubsection{Binary problems}

Let $\mathbf Z \in \mathbb{R}^{n\times d}$ and $y\in\{-1, 1\}^n$ be the training data. Logistic linear regression forecasts the conditional probability of class $1$ as 
\begin{equation}
    P(Y=1|\mathbf Z)=1 / \left( 1+\exp(-\boldsymbol \theta^\top \mathbf z)\right),
\end{equation}
where the vector $\boldsymbol \theta$ can be estimated by minimizing the loss function $\mathcal L(\boldsymbol \theta|\mathbf Z, \mathbf y) = \sum_i \log(1 + \exp(-y_i \boldsymbol\theta^\top \mathbf z))$. Since this loss function follows the Lipschitz condition, and a constant $ C$ limits the second derivative, Proposition~1 is still applicable~\cite{yu2022high} with $\boldsymbol a^k = \boldsymbol\theta^{(k-1)} - \frac{1}{C}\frac{\partial \mathcal L}{\partial \boldsymbol\theta}\left(\boldsymbol\theta^{(k-1)}\right) = \boldsymbol\theta^{(k-1)} + \frac{\mathbf z^\top y \exp(-y \mathbf z^\top \boldsymbol\theta^{(k-1)})}{1 + \exp(-y {\boldsymbol\theta^{(k-1)}}^\top \mathbf z)}$. As in the proposal, a B-spline transformer can be applied to the variables $\mathbf Z$, resulting in a new set $\tilde{\mathbf Z}$.

\subsubsection{Multi-class problems}

Let $\mathbf Z \in \mathbb{R}^{n\times d}$ and $y\in\{1,\dots, K\}$ be the training data. Logistic linear regression forecasts the conditional probability of a class $i$ as 
\begin{equation}
    P(Y=i|\mathbf Z) = \frac{\exp(\mathbf z^\top \boldsymbol\theta_i)}{\sum_{k=1}^K \exp(\mathbf z^\top \boldsymbol\theta_k)},
\end{equation}
where $\boldsymbol \theta_k$ can be obtained by minimizing the cross-entropy,
\begin{equation}
    CE(\boldsymbol\theta) = \frac{1}{T}\sum_{t=1}^T \sum_{k=1}^K \mathcal I_{(y_t=k)} \log(P(y_t=k|\mathbf Z_t)).
\end{equation}

\subsection{Seller Bids as a Function of the Coefficients}\label{sec:bids-times-coefficients}

In Section~\ref{sec:data-market-mechanism}, the $j$th seller bids $s_{j,k}$ for its $k$th variable, where the potential revenues are either zero or $s_{j,k}$. This bidding strategy makes the proposal a pool-based forecasting market. While this bidding approach is straightforward, sellers might prefer earning revenue related to their contributions, i.e., more aligned with a bilateral market. To address this, the concept from~\cite{trading22} can be merged with the cost-constrained restriction proposed in Section~\ref{sec:data-market-mechanism}. Here, seller $j$ has the choice to bid $s_{j,k}$ while the assigned value to receive becomes $s_{j,k} |\beta_{j,k}|$. Consequently, the market operator has to solve
\begin{subequations}
\label{eq:proposal-v2}
\begin{equation}
\label{eq:argmin_bid_constrained-v2}
\begin{aligned}
\hat{\boldsymbol{\Theta}}_i = & \underset{\boldsymbol\Theta_i=(\boldsymbol{\beta}^i, \boldsymbol{\eta}^i)}{\operatorname{argmin}} \, \mathcal L(\mathbf y_i, \sum \boldsymbol\Theta_i \tilde{\mathbf Z}_{i}) + \lambda \|\boldsymbol \Theta_i\|_1 \\ 
\end{aligned}
\end{equation}
\begin{equation}
\label{eq:bid-constraint-v2}
\begin{aligned}
\text{s.t. } &  \sum_{j=1}^N \left[\sum_{k=1}^{n_j}  s_{j,k}\sum_{m=1}^M|\beta_{k,j,m}^i| {+} \sum_{l=1}^L s_{j,n_j+l}\sum_{m=1}^M |\eta^i_{l,j,m}| \right] {\leq} b,
\end{aligned}
\end{equation}
\end{subequations}
for multiple bids $b \in \tilde{\boldsymbol b}$. This involves constructing a bid-gain table and defining the optimal price, as in~\eqref{eq:final-price}.

\section{Conclusion}\label{sec:conclusion}
This paper proposed a data market among RES data owners to improve RES forecasting accuracy by incentivizing data-sharing, especially among competitors. The solution enables data sellers to set prices for their data while granting buyers the flexibility to define either a maximum budget or a budget related to accuracy improvement when opting for collaborative forecasts instead of local ones. Subsequently, the market operator estimates a collaborative spline LASSO regression model, considering both feature prices and buyer budgets. This approach has two advantages: {i)}~it allows participants a simple bidding strategy; {ii)}~it uses a linear additive model (spline LASSO regression) that is computationally efficient and interpretable. The numerical results show that the proposed market can identify the most relevant resources in both low and high-buyer budget scenarios. The LASSO term removes redundant features, while budget constraints optimize feature allocation at minimal cost.

Future research could explore advanced buyer strategies and enhance the bidding process for sellers, considering the explanatory power of their variables (a key advantage of our interpretable solution) and possibly a minimum acceptable price. Additionally, future work could also explore meta-learning to leverage insights from other buyers, guiding hyperparameter selection~\cite{brazdil2022metalearning} and improving model performance and market computational efficiency. Furthermore, the proposed algorithmic solution should be integrated with digital technologies for data traceability and integrity, such as blockchain~\cite{Shang2022}. This combination provides a robust framework to prevent issues where data owners replicate or copy data, thereby maintaining trust in these mechanisms.

\appendix\label{sec:appendix}

\begin{proposition}[Pareto Efficiency]
By simplicity, consider $N$ data owners, each with a feature $X_i$ and price $s_i$. A data buyer wants to acquire forecasts of $Y_j$ with a budget that depends on the gain obtained with such forecasts, i.e., the problem is
\begin{equation}\label{eq:market_problem1}
\begin{aligned}
\arg_{\boldsymbol \beta}\max & \, \, \mathcal G(\mathcal L(\boldsymbol \beta)) 
\text{ s.t. }  Cost(\boldsymbol \beta)\leq \mathcal{VF}(\mathcal G(\mathcal L(\boldsymbol \beta))),    
\end{aligned}
\end{equation}
where $\boldsymbol{\beta}$ are linear regression coefficients, $\mathcal{L}(\boldsymbol{\beta})$ is the model's loss, $\mathcal{G}(\mathcal{L}(\boldsymbol{\beta}))$ is the gain, and $Cost(\boldsymbol{\beta})$ is the sum of allocated feature prices. Then, the allocation mechanism defined in~\eqref{eq:market_problem1} is Pareto efficient.
\end{proposition}

\begin{proof}
Since maximizing the gain is the same as minimizing the loss function, \eqref{eq:market_problem1} corresponds to:
\begin{equation}\label{eq:market_problem2}
\arg_{\beta}\min \mathcal L(\boldsymbol\beta)
\text{ s.t. } Cost(\boldsymbol\beta)\leq \mathcal{VF}(\mathcal G(\mathcal L(\boldsymbol\beta))).
\end{equation}
Assume $\boldsymbol\beta^*$ is the solution of~\eqref{eq:market_problem2} and it is not Pareto efficient. Then $\exists\boldsymbol\beta' \in \{\boldsymbol\beta \colon Cost(\beta)\leq \mathcal{VF}(\mathcal G(\mathcal L(\boldsymbol\beta)))\}$ such that:
$$\mathcal L(\boldsymbol\beta') > \mathcal L(\boldsymbol\beta^*) \quad \text{and} \quad \text{Cost}(\boldsymbol\beta') \leq \text{Cost}(\boldsymbol\beta^*).$$
Since $\boldsymbol\beta^*$ minimizes $\mathcal L(\boldsymbol\beta)$, this contradicts \(\boldsymbol\beta^*\) being the minimizer because $\beta'$ has a lower loss. Therefore, $\boldsymbol\beta^*$ is Pareto efficient.
\end{proof}

\begin{proposition}[Truthfullness]
The algorithm in~\eqref{eq:market_problem2} ensures that agents will report their true data.
\end{proposition}

\begin{proof}
Assume agent $i$ has data $X_i$, and he reports $X'_i$ to the data market operator. The data set all owners report is denoted by $\mathbf X' = (X'_1, \dots, X'_N)$. Then, the proposed mechanism estimates the parameters $(\beta^*_1(\boldsymbol{X}'),\dots, \beta^*_N(\boldsymbol{X}'))$ that minimize $\mathcal L(\beta_1, \dots, \beta_N$) s.t. $\sum_j s_j \beta_j\leq B$. Let $u_i(\beta_1, \dots, \beta_N, X_i)$ be the utility function  of agent $i$. Two scenarios are possible:
\begin{enumerate}
\item Agent $i$ reports original data $X_i$, and the mechanism allocates $(\beta^*_1(X_i, \boldsymbol{X}'_{-i}),\dots, \beta^*_N(X_i,\boldsymbol{X}'_{-i}))$
\item Agent $i$ reports $X'_i \neq X_i$, and the mechanism allocates $(\beta^*_1(X'_i, \boldsymbol{X}'_{-i}),\dots, \beta^*_N(X'_i,\boldsymbol{X}'_{-i}))$
\end{enumerate}
Since the utility is non-decreasing in the loss function, $u_i(\beta^*_1(X_i, \boldsymbol{X}'_{-i}),\dots, \beta^*_N(X_i,\boldsymbol{X}'_{-i})){\geq} \allowbreak u_i (\beta^*_1(X_i, \boldsymbol{X}'_{-i}),\dots, \beta^*_N(X'_i,\boldsymbol{X}'_{-i})).$
\end{proof}

\ifCLASSOPTIONcaptionsoff
  \newpage
\fi

\bibliographystyle{IEEEtran}
\bibliography{bibliography}

\begin{IEEEbiographynophoto}{Carla Gon\c{c}alves} holds a Ph.D. in Applied Mathematics from the University of Porto. Currently, she is researching forecasting methods and data-sharing incentives at the Center for Power and Energy Systems, INESC TEC.
\end{IEEEbiographynophoto}

\begin{IEEEbiographynophoto}{Ricardo J. Bessa} received the Licenciado degree in Electrical and Computer Engineering, the M.Sc. degree in Data Analysis and Decision Support Systems, and the Ph.D. degree in Sustainable Energy Systems from the  University of  Porto.  He is the coordinator of the Center for Power and Energy Systems, INESC TEC. His research interests include renewable energy,  energy analytics,  smart grids,  and electricity markets. Associate Editor of the Journal of Modern Power Systems and Clean Energy, received the Energy Systems Integration Group (ESIG) Excellence Award in 2022.
\end{IEEEbiographynophoto}

\begin{IEEEbiographynophoto}{Tiago Teixeira} holds a B.Sc. in Physics and an M.Sc. in Data Science from the University of Porto. His interests include data analytics and forecasting.
\end{IEEEbiographynophoto}

\begin{IEEEbiographynophoto}{Jo\~{a}o Vinagre} received an M.Sc. in Network and Systems Engineering and a Ph.D. in Computer Science from the University of Porto. His research interests include collaborative AI applications and human-AI interaction, focusing on recommender systems and decentralized machine learning. He is a scientific project officer at the Joint Research Centre of the European Commission, working in the European Centre for Algorithmic Transparency team.
\end{IEEEbiographynophoto}

\end{document}